%% file: main.tex
\PassOptionsToPackage{hyperfootnotes=false}{hyperref}
\documentclass{lab}
\usepackage[utf8]{inputenc}
\usepackage[T1]{fontenc}
\usepackage{charter}
\usepackage{xurl}
\usepackage{placeins}

\input{configuration/preamble.tex}

\input{configuration/apex_layout.tex}
\definecolor{color1}{HTML}{406058}
\definecolor{color2}{HTML}{78A0C8}
\definecolor{color3}{HTML}{3850A0}
\definecolor{color4}{HTML}{6860A8}
\definecolor{color5}{HTML}{A03850}
\definecolor{color6}{HTML}{AE7694}
\definecolor{color7}{HTML}{8488B4}
\hypersetup{
    pdftitle={Few-Step Streaming Video Generation via Condition Shifting},
    pdfauthor={Deyuan Liu, Jungang Li, Kaiwen Zheng, Kai Jiang, Jintao Chen, Yaole Wang, Ming Lin, Tao Lin, Jianfei Chen, Jun Zhu, Jintao Zhang},
    colorlinks=true,
    linkcolor=black,
    citecolor=color1,
    urlcolor=color2,
    pdfborder={0 0 0}
}

\title{Few-Step Streaming Video\\Generation via Condition Shifting}
\author{{\fontsize{10.2}{13}\selectfont
Deyuan Liu$^{1,2,3,*}$ \quad
Jungang Li$^{1,2,*}$ \quad
Kaiwen Zheng$^{1,2,*}$ \quad
Kai Jiang$^{1,2,*}$ \quad
Jintao Chen$^{1,2}$ \\[1mm]
Yaole Wang$^{1,2}$ \quad
Ming Lin$^{1,2}$ \quad
Tao Lin$^{3,\ddagger}$ \quad
Jianfei Chen$^{1,2,\ddagger}$ \quad
Jun Zhu$^{1,2,\ddagger}$ \quad
Jintao Zhang$^{1,2,\ddagger}$}}
\affiliation{$^{1}$Tsinghua University \quad $^{2}$Shengshu Technology \quad $^{3}$Westlake University}
\input{sections/0_abstract.tex}
\begin{document}
\maketitle
\vspace{-6mm}
\begingroup
\renewcommand{\thefootnote}{}
\footnotetext{*Co-first authorship. $^{\ddagger}$Corresponding authors.}
\endgroup
\setcounter{footnote}{0}
\input{figures/1_case_gallery.tex}

\fontsize{10}{11.6}\selectfont
\setlength{\parskip}{0.7pt plus0.5pt minus0.5pt}
\setlength{\abovedisplayskip}{6pt plus2pt minus2pt}
\setlength{\belowdisplayskip}{6pt plus2pt minus2pt}
\setlength{\abovedisplayshortskip}{3pt plus1pt minus1pt}
\setlength{\belowdisplayshortskip}{5pt plus1pt minus1pt}
\setlength{\textfloatsep}{8pt plus2pt minus1pt}
\setlength{\floatsep}{8pt plus2pt minus1pt}
\setlength{\intextsep}{7pt plus2pt minus1pt}
\captionsetup{skip=4pt}

\input{sections/1_introduction.tex}
\input{sections/2_preliminaries.tex}
\input{sections/3_method.tex}
\input{sections/4_evaluation.tex}

\FloatBarrier
\input{sections/5_conclusion.tex}
\normalsize
\captionsetup{skip=10pt}

\clearpage

\bibliography{configuration/bibstrings,resources/reference}
\bibliographystyle{configuration/iclr2027_conference}

\clearpage
\appendix

\onecolumn
{
    \hypersetup{linkcolor=black}
    \parskip=0em
    \renewcommand{\contentsname}{Contents}
    \setcounter{tocdepth}{3}
    \setlength{\cftbeforesecskip}{5pt}
    \tableofcontents
}
\clearpage

\input{sections/6_related_work.tex}

\input{sections/7_theoretical_analysis.tex}
\clearpage
\input{sections/8_training_data.tex}

\input{sections/9_evaluation_protocol.tex}

\end{document}

%% file: configuration/preamble.tex
\usepackage{amsmath,amssymb,amsthm}

\providecommand{\R}{\mathbb{R}} %
\providecommand{\EEb}[2]{{\mathbb E}_{#1}\left[#2\right] } %

\providecommand{\cc}{\mathbf{c}}
\providecommand{\dd}{\mathbf{d}}

\providecommand{\hh}{\mathbf{h}}

\renewcommand{\ss}{\mathbf{s}}

\providecommand{\uu}{\mathbf{u}}
\providecommand{\vv}{\mathbf{v}}

\providecommand{\xx}{\mathbf{x}}
\providecommand{\yy}{\mathbf{y}}
\providecommand{\zz}{\mathbf{z}}

\providecommand{\mI}{\mathbf{I}}

\providecommand{\mf}{\boldsymbol{f}}

\providecommand{\mmF}{\boldsymbol{F}}

\providecommand{\mmG}{\boldsymbol{G}}

\providecommand{\mepsilon}{\boldsymbol{\epsilon}}

\providecommand{\mdelta}{\boldsymbol{\delta}}

\providecommand{\cG}{\mathcal{G}}

\providecommand{\cL}{\mathcal{L}}

\providecommand{\cN}{\mathcal{N}}

\providecommand{\cU}{\mathcal{U}}

\newenvironment{talign*}
{\csname align*\endcsname}
{\endalign}

\usepackage{nicefrac}
\usepackage{algorithm}
\usepackage{algorithmic}
\usepackage{graphicx}
\usepackage{subcaption}
\usepackage[flushleft]{threeparttable}
\usepackage{float}
\usepackage{multirow}
\usepackage{xspace}
\usepackage{natbib}
\usepackage{enumitem}
\usepackage{caption}
\usepackage{autobreak}
\usepackage{sidecap}
\usepackage{wrapfig}
\usepackage{bbding}
\usepackage[toc, page, header]{appendix}
\usepackage{tikz}
\usetikzlibrary{arrows.meta,positioning,calc}
\usepackage{pifont}
\usepackage{mdframed}
\usepackage{tcolorbox}
\tcbuselibrary{skins,breakable}
\usepackage{colortbl}
\usepackage{mathrsfs}
\usepackage{footmisc}

\newsavebox{\apextablemeasure}
\newcount\apextablepaddingcount
\newcommand{\apexfilledtable}[3]{%
  \begingroup
  \setlength{\tabcolsep}{0pt}%
  \sbox{\apextablemeasure}{\begin{tabular}{#2}#3\end{tabular}}%
  \apextablepaddingcount=\numexpr2*#1\relax
  \setlength{\tabcolsep}{\dimexpr(\linewidth-\wd\apextablemeasure)/\apextablepaddingcount\relax}%
  \begin{tabular}{#2}#3\end{tabular}%
  \endgroup
}

\definecolor{coral}{RGB}{255,127,80}
\definecolor{darkgreen}{RGB}{0,100,0}
\definecolor{darkyellow}{RGB}{204,153,0}
\definecolor{salmon}{RGB}{250,128,114}
\definecolor{champagne}{RGB}{247,231,206}
\definecolor{kleinblue}{RGB}{0,47,167}
\definecolor{bauhiniapurple}{RGB}{193,84,193}
\definecolor{darkred}{RGB}{150,0,0}

\newcommand{\darkredtext}[1]{{\color{darkred}#1}}

\newcommand{\thmref}[1]{\hyperref[#1]{\darkredtext{Thm.~\ref*{#1}}}}
\newcommand{\defref}[1]{\hyperref[#1]{\darkredtext{Def.~\ref*{#1}}}}
\newcommand{\propref}[1]{\hyperref[#1]{\darkredtext{Prop.~\ref*{#1}}}}
\newcommand{\assumpref}[1]{\hyperref[#1]{\darkredtext{Assump.~\ref*{#1}}}}
\newcommand{\remarkref}[1]{\hyperref[#1]{\darkredtext{Rem.~\ref*{#1}}}}
\newcommand{\hypref}[1]{\hyperref[#1]{\darkredtext{Hyp.~\ref*{#1}}}}
\newcommand{\conjref}[1]{\hyperref[#1]{\darkredtext{Conj.~\ref*{#1}}}}
\newcommand{\lemref}[1]{\hyperref[#1]{\darkredtext{Lem.~\ref*{#1}}}}
\newcommand{\corref}[1]{\hyperref[#1]{\darkredtext{Cor.~\ref*{#1}}}}
\newcommand{\noteref}[1]{\hyperref[#1]{\darkredtext{Nota.~\ref*{#1}}}}
\newcommand{\claimref}[1]{\hyperref[#1]{\darkredtext{Clm.~\ref*{#1}}}}
\newcommand{\obsref}[1]{\hyperref[#1]{\darkredtext{Obs.~\ref*{#1}}}}

\newcommand{\transparentteal}[1]{%
  \tikz[baseline=(X.base)] \node[fill=teal, fill opacity=0.1, text opacity=1, inner sep=2pt, outer sep=0pt] (X) {#1};%
}
\newcommand{\figref}[1]{\hyperref[#1]{\transparentteal{Figure~\ref*{#1}}}}

\newcommand{\transparentdarkgreen}[1]{%
  \tikz[baseline=(X.base)] \node[fill=darkgreen, fill opacity=0.1, text opacity=1, inner sep=2pt, outer sep=0pt] (X) {#1};%
}
\newcommand{\tabref}[1]{\hyperref[#1]{\transparentdarkgreen{Table~\ref*{#1}}}}

\newcommand{\transparentdarkyellow}[1]{%
  \tikz[baseline=(X.base)] \node[fill=darkyellow, fill opacity=0.1, text opacity=1, inner sep=2pt, outer sep=0pt] (X) {#1};%
}
\newcommand{\secref}[1]{\hyperref[#1]{\transparentdarkyellow{Section~\ref*{#1}}}}

\newcommand{\transparentcoral}[1]{%
  \tikz[baseline=(X.base)] \node[fill=coral, fill opacity=0.1, text opacity=1, inner sep=2pt, outer sep=0pt] (X) {#1};%
}
\newcommand{\appref}[1]{\hyperref[#1]{\transparentcoral{Appendix~\ref*{#1}}}}

\newcommand{\transparentkleinblue}[1]{%
  \tikz[baseline=(X.base)] \node[fill=kleinblue, fill opacity=0.1, text opacity=1, inner sep=2pt, outer sep=0pt] (X) {#1};%
}
\newcommand{\algref}[1]{\hyperref[#1]{\transparentkleinblue{Algorithm~\ref*{#1}}}}

\newcommand{\transparentbauhiniapurple}[1]{%
  \tikz[baseline=(X.base)] \node[fill=bauhiniapurple, fill opacity=0.1, text opacity=1, inner sep=2pt, outer sep=0pt] (X) {#1};%
}
\newcommand{\equref}[1]{\hyperref[#1]{\transparentbauhiniapurple{Equation~\ref*{#1}}}}

\newtheoremstyle{custom}
{1pt} 
{1pt} 
{\itshape} 
{} 
{\bfseries} 
{} 
{ } 
{\thmname{#1} \thmnumber{#2} \thmnote{(#3)} . } 

\theoremstyle{custom}

\newtheorem{innerdefinition}{Definition}
\newtheorem{innerproposition}{Proposition}
\newtheorem{innerassumption}{Assumption}
\newtheorem{innerremark}{Remark}
\newtheorem{innertheorem}{Theorem}
\newtheorem{innerhypothesis}{Hypothesis}
\newtheorem{innerconjecture}{Conjecture}
\newtheorem{innerlemma}{Lemma}
\newtheorem{innercorollary}{Corollary}
\newtheorem{innerexample}{Example}
\newtheorem{innernotation}{Notation}
\newtheorem{innerclaim}{Claim}
\newtheorem{innerproblem}{Problem}

\newtheorem{innerobservation}{Observation}

\tcbset{
  theoremframebase/.style={
    enhanced,
    breakable,
    sharp corners,
    boxrule=0pt,
    leftrule=2pt,
    top=6pt,
    bottom=5pt,
    left=5pt,
    right=5pt,
    before skip=4pt,
    after skip=4pt,
  }
}

\newtcolorbox{definitionframe}{theoremframebase, colback=gray!10, colframe=gray!75!black}
\newtcolorbox{propositionframe}{theoremframebase, colback=blue!8, colframe=blue!70!black}
\newtcolorbox{assumptionframe}{theoremframebase, colback=green!8, colframe=green!55!black}
\newtcolorbox{remarkframe}{theoremframebase, colback=yellow!12, colframe=yellow!60!black}
\newtcolorbox{theoremframe}{theoremframebase, colback=red!7, colframe=red!70!black}
\newtcolorbox{hypothesisframe}{theoremframebase, colback=purple!7, colframe=purple!70!black}
\newtcolorbox{conjectureframe}{theoremframebase, colback=orange!8, colframe=orange!70!black}
\newtcolorbox{lemmaframe}{theoremframebase, colback=cyan!8, colframe=cyan!70!black}
\newtcolorbox{corollaryframe}{theoremframebase, colback=magenta!7, colframe=magenta!70!black}
\newtcolorbox{exampleframe}{theoremframebase, colback=lime!8, colframe=lime!45!black}
\newtcolorbox{notationframe}{theoremframebase, colback=pink!7, colframe=pink!70!black}
\newtcolorbox{claimframe}{theoremframebase, colback=violet!7, colframe=violet!70!black}
\newtcolorbox{problemframe}{theoremframebase, colback=salmon!9, colframe=salmon!70!black}
\newtcolorbox{observationframe}{theoremframebase, colback=purple!5, colframe=purple!55!black}

\newenvironment{proposition}
{\begin{propositionframe}\begin{innerproposition}}
{\end{innerproposition}\end{propositionframe}}

\newenvironment{remark}
{\begin{remarkframe}\begin{innerremark}}
{\end{innerremark}\end{remarkframe}}

\newenvironment{theorem}
{\begin{theoremframe}\begin{innertheorem}}
{\end{innertheorem}\end{theoremframe}}

\newenvironment{corollary}
{\begin{corollaryframe}\begin{innercorollary}}
{\end{innercorollary}\end{corollaryframe}}

\usepackage[textwidth=3.5cm]{todonotes}

\newcommand{\APEX}{\textsc{APEX}\xspace}

\newcommand{\sg}{\mathop{{\color{red}\operatorname{sg}}}\nolimits}

\usepackage{makecell}

\usepackage{bm}

\definecolor{ForestGreen}{RGB}{34,139,34}
\definecolor{OrangeRed}{RGB}{255,69,0}



%% file: configuration/apex_layout.tex
\usepackage{needspace}
\DeclareTextFontCommand{\textbf}{\bfseries}
\setcitestyle{authoryear,round,citesep={;},aysep={,},yysep={;}}
\makeatletter
\def\NAT@sort{0}
\def\NAT@cmprs{0}
\makeatother
\BeforeBeginEnvironment{proposition}{\Needspace{12\baselineskip}}

\usepackage{fix-cm}
\AtBeginDocument{%
  \ifcsname T1+bch\endcsname
    \DeclareFontShape{T1}{bch}{m}{scit}{<->ssub*bch/m/sc}{}%
  \fi
}

\titlespacing*{\section}{0pt}{8pt plus2pt minus1pt}{5pt}
\titlespacing*{\subsection}{0pt}{6pt plus2pt minus1pt}{3pt}
\titlespacing*{\paragraph}{0pt}{5pt plus1pt minus1pt}{0.7em}

\definecolor{apextablehead}{HTML}{E7EFF5}
\definecolor{apextableours}{HTML}{D7E6F0}

%% file: sections/0_abstract.tex
\abstract{
  Few-step sampling reduces the cost of streaming video generation.
  Distribution Matching Distillation (DMD) supervises few-step generators with a separate fake-score network, which adds training memory and computation and must track the evolving generator.
  We extend \textcolor{color3}{\textbf{\APEX}} to streaming video by using condition shifting to estimate the velocity field of the generated distribution within the generator.
  The original text condition selects generation, while a shifted condition selects fake estimation using the same parameters.
  This removes the independent fake network while retaining S2-Avatar's frozen reference teacher, Self-Replay Forcing, and perceptual regularization.
  We evaluate APEX on 180-second videos using three backbone evaluations per block and a one-step spatial Refiner, with a 70-case human study and a 320-case StreamAV-Bench comparison.
  Compared with S2-Avatar, APEX achieves higher video instruction fulfillment on the Progressive track (3.856 versus 3.831) and higher visual aesthetics across both tracks (0.641898 versus 0.632606).
  Human raters favor APEX most strongly on instruction following, with $+30.0\%$ net preference, followed by consistency and video quality at $+4.3\%$ each.
  Motion quality favors S2-Avatar, audiovisual synchronization is predominantly tied, and usability is unchanged.
  Separate measurements under matched training settings show $4.4\%$ lower peak allocated GPU memory per GPU and $9.2\%$ shorter mean iteration time than DMD.
}

%% file: figures/1_case_gallery.tex
\begin{figure}[p]
  \centering
  \includegraphics[width=\linewidth]{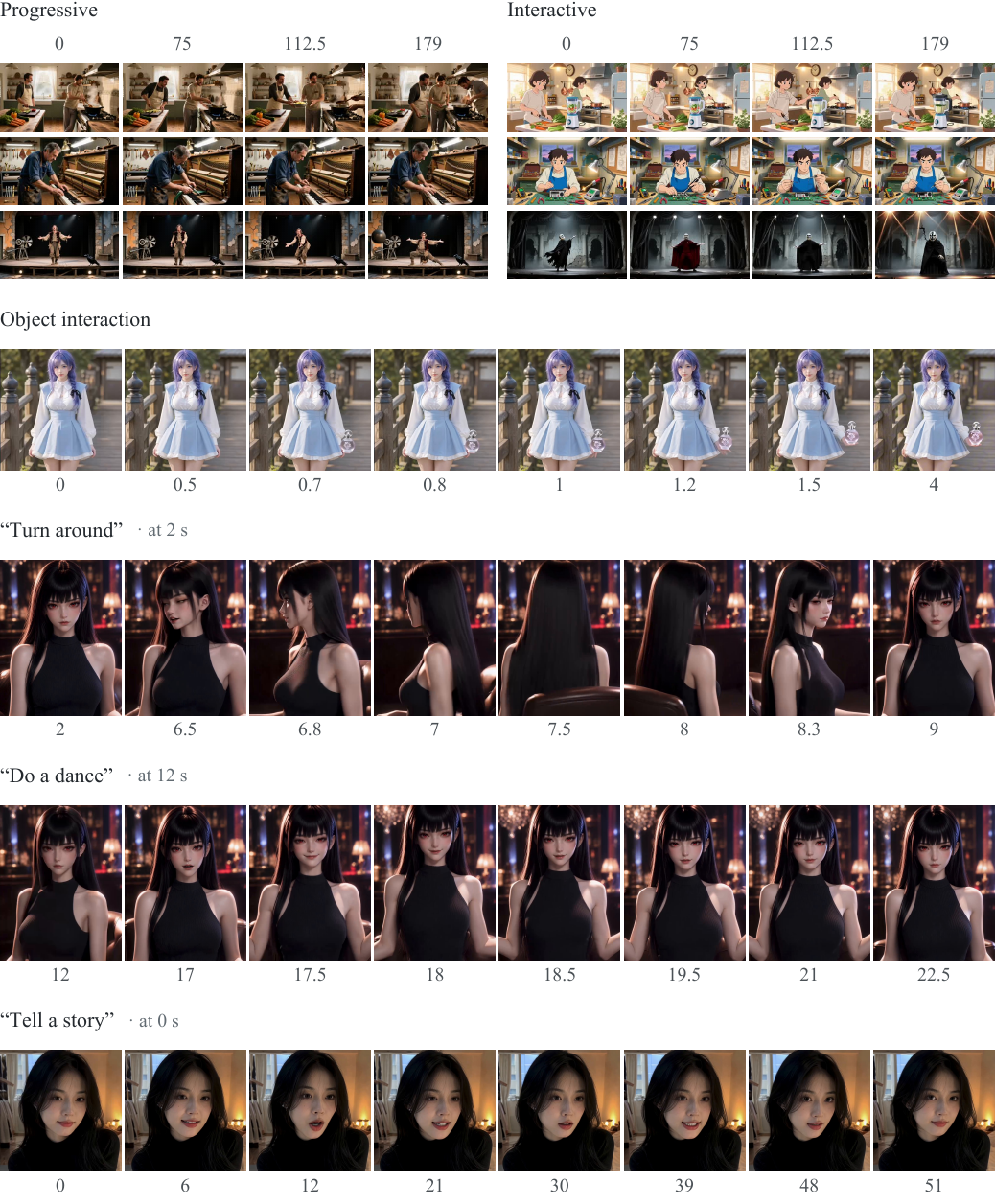}
  \caption{\textbf{APEX streaming generation and interaction.} Top: Progressive (left) and Interactive (right) StreamAV-Bench samples (3 NFE, $2\times$ Refiner). The lower four rows show object interaction, turning, dancing, and storytelling, with each sequence read from left to right. Quoted user prompts are translated from Chinese, with rounded onset times. All timestamps are in seconds. Demo times are measured relative to the start of each recording.}
  \label{fig:case_gallery}
\end{figure}

%% file: sections/1_introduction.tex
\section{Introduction}
Diffusion and flow models produce high-quality videos through repeated network evaluations~\citep{videoworldsimulators2024,veo,wan2025wan,seedance2026seedance}.
Streaming generation incurs this cost for every new block, making few-step sampling particularly valuable for long sequences.
Distillation reduces the number of evaluations~\citep{song2023consistency,salimans2022progressive,zheng2025rcm}, while the resulting generator must preserve visual quality, maintain characters and scenes, and respond to changing instructions.
\figref{fig:case_gallery} shows APEX video examples of object interaction and storytelling.

Distribution Matching Distillation (DMD) supervises generated samples through the difference between reference and generated-distribution scores~\citep{yin2024one,yin2024improved}.
A frozen pretrained teacher supplies the reference estimate, while a separate fake-score network learns from current generator outputs.
For large video backbones, maintaining this auxiliary network adds training memory and computation.
Its estimate must also track the generator as the output distribution changes, requiring coordination of their optimization settings and updates.
This motivates learning the fake estimate within the generator.

TwinFlow~\citep{cheng2026twinflow} and the original \textcolor{color3}{\textbf{\APEX}}~\citep{liu2026self} show how generation and distribution estimation can share a network.
We extend APEX's condition shifting to streaming video: the original text condition selects generation, and its shifted version selects a fake estimator trained on outputs for the same prompt.
The two roles share model parameters and, by default, optimizer states, while a separate frozen pretrained teacher provides the same reference guidance as in S2-Avatar.

Streaming video requires the estimator to track outputs generated with a causal history.
We integrate condition shifting into S2-Avatar's Self-Replay Forcing (SRF), which trains on replayed segments of the model's own generated sequences~\citep{zhang2026vidu}.
The shared fake estimator is fitted to these replay outputs, and its clean-latent predictions are compared with those of the frozen teacher to construct a detached correction for the generator.
We retain S2's replay procedure and perceptual regularization, as described in \secref{sec:video_method}.

We evaluate APEX on 180-second videos through a 70-case human study and all 320 StreamAV-Bench scenarios.
Relative to S2-Avatar, APEX receives positive net human preference for instruction following, consistency, and video quality, and achieves higher video instruction-fulfillment scores and visual aesthetics on both benchmark tracks.
A separate measurement under matched training settings shows $4.4\%$ lower peak allocated GPU memory and $9.2\%$ shorter mean iteration time than DMD.
Instruction-following preference also remains positive at 2 backbone NFE.
\secref{sec:exp} also reports audiovisual synchronization, motion quality, and usability.

The paper makes three contributions to few-step streaming video generation:
\begin{enumerate}[leftmargin=*,nosep]
\item \textbf{Shared-network estimation through condition shifting.} We learn the fake velocity within the generator while retaining a frozen reference teacher and removing the independent fake network.
\item \textbf{Integration with Self-Replay Forcing.} We apply condition-shifted estimation to the outputs of S2's Self-Replay Forcing, retaining its replay procedure and perceptual regularization.
\item \textbf{Video quality and training efficiency.} We evaluate 180-second generation on 70 paired cases and all 320 StreamAV-Bench scenarios, alongside training-cost and sampling comparisons.
\end{enumerate}

%% file: sections/2_preliminaries.tex
\section{Preliminaries}
\label{sec:preliminaries}
We use $\xx$ for a data latent, $\zz$ for Gaussian noise, $\cc$ for projected text, and $\mmF_\theta$ for a velocity predictor. The operator $\sg$ denotes stop-gradient, and \secref{sec:s2_training} defines the video context.

\paragraph{Continuous Generative Models.}
\label{sec:continuous-path}
Conditional diffusion~\citep{ho2020denoising,song2020score} and flow matching~\citep{lipman2022flow} transport Gaussian noise $\zz\sim p_z=\cN(\mathbf{0},\mI)$ to data $\xx\sim p_r(\cdot\mid\cc)$.
For $t\in[0,1]$, we use the linear interpolation path of optimal transport (OT) flow matching:
\begin{equation}
  \xx_t = t\zz + (1-t)\xx \,.
  \label{eq:xt_interpolant}
\end{equation}
Sampling transports noise $\xx_1=\zz$ to data $\xx_0=\xx$ along the probability-flow ODE:
\begin{equation}
  \frac{\mathrm{d}\xx_t}{\mathrm{d}t}=\vv(\xx_t,t,\cc) \,.
  \label{eq:pf_ode}
\end{equation}
Flow matching learns this velocity field by regressing to the known target $\uu_r(\xx,\zz):=\zz-\xx$:
\begin{equation}
  \cL_{\mathrm{FM}}(\theta)=\EEb{\xx,\,\cc,\,\zz,\,t}{\|\mmF_\theta(\xx_t,t,\cc)-\uu_r(\xx,\zz)\|^2} \,.
  \label{eq:fm_loss}
\end{equation}
The expectation uses paired data $(\xx,\cc)$, independent $\zz\sim p_z$, and $t\sim\cU(0,1)$.
The population minimizer is the conditional mean velocity $\vv_r^*(\xx_t,t,\cc)=\mathbb E[\zz-\xx\mid\xx_t,t,\cc]$.
Sampling integrates from $t=1$ to $t=0$ with numerical ODE solvers~\citep{karras2022edm}, such as the first-order Euler method.

\paragraph{Few-Step Generation.}
\label{sec:fewstep}
Consistency and interval-mapping methods support few-step sampling~\citep{song2023consistency,wang2024phased,zheng2025rcm,sun2025anystep,geng2025mean}.
The one-step Euler map is $\mmG_\theta(\zz,\cc)=\zz-\mmF_\theta(\zz,1,\cc)$, with no destination-time input to the velocity predictor.

\paragraph{Distribution Matching.}
For a fixed condition, let $p_\theta$ be the distribution of $\mmG_\theta(\zz,\cc)$ and $p_r$ the reference distribution.
Their noised marginals at time $t$ have scores $\ss_{i,t}(\yy):=\nabla_\yy\log p_{i,t}(\yy)$ for $i\in\{r,\theta\}$.
DMD~\citep{yin2024one} estimates the reverse-KL gradient using their score difference at the same noised generated sample.
It uses a pretrained reference score and an auxiliary fake-score network trained on current outputs.
\appref{app:score_velocity} expresses this score difference through marginal velocities under the OT path.

TwinFlow~\citep{cheng2026twinflow} learns both velocities within one model, using positive time for the real trajectory and negative time for the fake trajectory.
APEX selects the fake estimator through the condition input. Our video implementation shares the generator and fake estimator and retains S2's frozen teacher for the reference prediction.
\appref{sec:gan-dynamics} relates the GAN gradient to a weighted score difference.

\subsection{Autoregressive Video Generation and Self-Replay Forcing}
\label{sec:s2_training}
We express S2-Avatar's causal predictor and Self-Replay Forcing procedure~\citep{zhang2026vidu} in APEX notation.
For block $i$, let $\xx^i$ denote the video latent and $\cc^i$ its text representation.
The context $\hh^i$ contains the reference image, the preceding video states, and any accompanying audio.
The diffusion time $t_i\in(0,1]$ is its noise level, with larger values indicating more noise.
The clean prediction is
\begin{equation}
\widehat{\xx}_0^i
=\xx_{t_i}^i-t_i\mmF_\theta(\xx_{t_i}^i,t_i,\cc^i\mid\hh^i).
\label{eq:s2_causal}
\end{equation}
A block-wise causal mask lets each block attend to its own tokens, conditions, and preceding video blocks.

\paragraph{Self-Replay Forcing.}
Following S2~\citep{zhang2026vidu}, we generate a detached autoregressive rollout~\citep{huang2026self} and re-noise a sampled segment with diffusion forcing (DF)~\citep{DF}.
One differentiable causal pass processes this segment, with each block reading preceding blocks at their replay noise levels.
This allows cross-block gradients within the segment while keeping the external history and original rollout detached.
We retain S2's teacher and perceptual regularization~\citep{vidus1} and replace DMD's fake estimator~\citep{yin2024one,yin2024improved} through condition shifting in \secref{sec:video_method}.

%% file: sections/3_method.tex
\section{Condition Shifting for Streaming Video Generation}
\label{sec:apex}
\label{sec:video_method}
We apply APEX's condition shifting~\citep{liu2026self} to the fake estimator in S2's Self-Replay Forcing framework.
The generator and fake estimator share parameters $\theta$, with a frozen teacher $\mmF_\phi$ providing S2's reference prediction and guidance.
\figref{fig:apex_method} compares DMD and APEX during training.

\subsection{Condition Shifting on Generated Video}
\label{subsec:key_concepts}
\label{subsec:fake_branch}
The network generates under $\cc^i$ and estimates the fake velocity from these outputs under $\cc_{\mathrm{shift}}^i=-\cc^i$.
We negate projected text before the transformer blocks, with positive diffusion times in both calls.

The $N$ evaluated blocks $\xx_\theta^{1:N}$ include replay outputs and any detached prefix from \secref{sec:s2_training}.
We detach them and re-noise at a scoring time $t'\in(0,1)$ with fresh Gaussian noise $\mepsilon'\sim p_z$:
\begin{equation}
\xx_\theta^{1:N,\sg}:=\sg(\xx_\theta^{1:N}),\qquad
\xx_{\theta,t'}^{1:N,\sg}:=(1-t')\xx_\theta^{1:N,\sg}+t'\mepsilon'.
\label{eq:video_model_path}
\end{equation}
We draw a single scoring time $t'$ for all generated blocks in a sequence, separately from the replay times $t_i$ used to produce the clean predictions.
With reference-image and audio conditions fixed and omitted from the notation, the shifted-condition network predicts the fake velocity from the re-noised video latents:
\begin{equation}
\widehat\vv_{\theta,t'}^{1:N}
:=\mmF_\theta(\xx_{\theta,t'}^{1:N,\sg},t',\cc_{\mathrm{shift}}^{1:N}).
\label{eq:video_fake_prediction}
\end{equation}
To learn the velocity field of the generated samples, we use the following flow-matching loss with causal attention over the current and preceding noisy blocks, without an additional clean-video stream:
\begin{equation}
\cL_{\mathrm{fake}}(\theta)
=\mathbb E\!\left[\operatorname{MSE}\!\left(
\widehat\vv_{\theta,t'}^{1:N},\mepsilon'-\xx_\theta^{1:N,\sg}
\right)\right].
\label{eq:video_fake}
\end{equation}
Here MSE is the mean squared error with S2's video weighting and averaging.
Detaching the generated sequence and regression target restricts this update to the shifted-condition prediction.

\input{figures/2_apex_method.tex}

\subsection{A Detached Correction from the Velocity Difference}
\label{subsec:gan_connection}
We form the generator correction from the frozen teacher's $\widehat\xx_{0,r}^{1:N}$ and fake clean-latent predictions at the same noisy video and scoring time, retaining S2's conditioning, guidance, and DMD target~\citep{zhang2026vidu,yin2024one,yin2024improved}.
Here $r$ marks the teacher prediction and $\theta$ the fake estimate.
\begin{equation}
\widehat\xx_{0,\theta}^{i}
=\xx_{\theta,t'}^{i,\sg}-t'\widehat\vv_{\theta,t'}^{i},\qquad
\widehat\vv_{r,t'}^{i}
:=\frac{\xx_{\theta,t'}^{i,\sg}-\widehat\xx_{0,r}^{i}}{t'}.
\label{eq:video_branches}
\end{equation}
Their clean-latent difference is proportional to the velocity difference at the same scoring time:
\begin{equation}
\widehat\xx_{0,r}^{i}-\widehat\xx_{0,\theta}^{i}
=t'\widehat\dd_{t'}^i,\qquad
\widehat\dd_{t'}^i:=\widehat\vv_{\theta,t'}^i-\widehat\vv_{r,t'}^i.
\label{eq:video_endpoint_difference}
\end{equation}
We normalize this clean-latent difference as in \appref{app:apex_velocity_correction} to obtain $\widetilde\dd_{t'}^{1:N}$ and add it to the current output, giving the generator a detached regression target for the replay predictions:
\begin{equation}
\cL_{\mathrm{score}}(\theta)
=\mathbb E\!\left[\operatorname{MSE}\!\left(\xx_\theta^{1:N},\sg\!\left(\xx_\theta^{1:N}+\widetilde\dd_{t'}^{1:N}\right)\right)\right].
\label{eq:video_correction}
\end{equation}
We hold the target fixed, including estimator predictions and normalization.
Gradients therefore pass through replay outputs and their connected causal representations within the active segment.

Under the OT path, velocity estimates induce score estimates satisfying
\begin{equation}
\widehat\dd_{t'}^i
=-\frac{t'}{1-t'}\left(\widehat\ss_{\theta,t'}^i-\widehat\ss_{r,t'}^i\right).
\label{eq:apex_score_difference}
\end{equation}
Interpreting these induced estimates as true conditional scores requires calibrated fields under common conditioning.
\appref{app:apex_velocity_correction} derives the gradient of \equref{eq:video_correction} and discusses its score interpretation.

\subsection{Shared-Parameter Updates}
\label{subsec:complete_obj}
Following S2, the generator objective combines distribution matching and perceptual regularization:
\begin{equation}
\cL(\theta)=\cL_{\mathrm{score}}(\theta)+\cL_{\mathrm{perc}}(\theta).
\label{eq:video_objective}
\end{equation}
Generator updates alternate with fake estimation in \equref{eq:video_fake} on shared parameters $\theta$, retaining S2's perceptual regularization~\citep{zhang2026vidu,vidus1} and keeping $\phi$ frozen.
By default, these updates share AdamW states, with separate states evaluated in \secref{sec:video_additional}.
At inference, the original projected text selects generation with the chosen few-step schedule, followed by the common one-step spatial Refiner.

%% file: figures/2_apex_method.tex
\begin{figure}[!t]
  \centering
  \includegraphics[width=\linewidth]{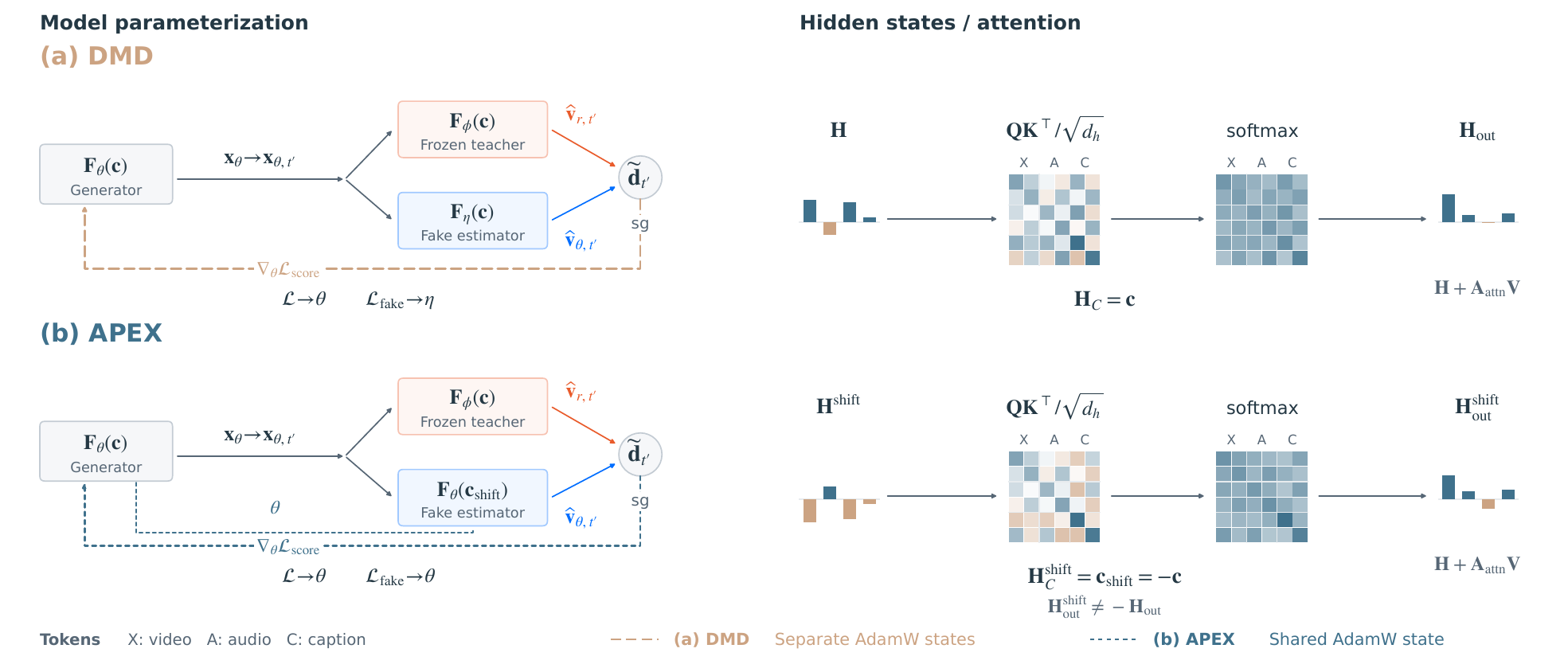}
  \caption{\textbf{DMD and APEX in Self-Replay Forcing.} DMD uses independent fake parameters $\eta$; APEX shares generator and fake parameters $\theta$ and, by default, AdamW states. Teacher parameters $\phi$ are frozen. Left: $\xx_\theta$ is the rollout/replay output, $t'$ the scoring time, $\widetilde\dd_{t'}$ the normalized correction, and $\sg$ stop-gradient. The correction contributes $\nabla_\theta\cL_{\mathrm{score}}$; $\cL$ is the full generator objective. The fake-velocity subscript $\theta$ identifies the generated distribution. Right: $\mathbf H$ and $\mathbf H_{\mathrm{out}}$ denote input and output hidden states; $\mathbf H_C$ contains projected text tokens $\cc$. Within each row, $\mathbf Q,\mathbf K,\mathbf V$ are queries, keys, and values, and $\mathbf A_{\mathrm{attn}}=\operatorname{softmax}(\mathbf Q\mathbf K^\top/\sqrt{d_h})$, with head dimension $d_h$. Superscript $\mathrm{shift}$ marks states under $\cc_{\mathrm{shift}}=-\cc$. The single-head example uses fixed weights and omits the causal mask and other transformer operations; values are schematic.}
  \label{fig:apex_method}
\end{figure}

%% file: sections/4_evaluation.tex
\section{Evaluation}
\label{sec:exp}
We evaluate 180-second streaming generation on ten StreamAV-Bench metrics and five human-preference dimensions, complemented by absolute ratings, usability, and training-cost and sampling comparisons.

\subsection{Setting}
\label{sec:video_setup}

\paragraph{Training data and model.}
We reuse S2-Avatar's complete training data and preparation pipeline~\citep{zhang2026vidu}, including filtering and annotation, for livestreams, talking heads, film and television, solo dance, and 2D/3D animation, as summarized in \appref{sec:data}. APEX uses its causal audiovisual backbone and Self-Replay Forcing, sharing parameters and, by default, AdamW states for generator and fake updates.

\paragraph{Inference.}
Unless otherwise specified, both models use 3 backbone denoising function evaluations (NFE) per generated block, excluding KV-cache encoding and updates. The common one-step $2\times$ spatial Refiner is counted separately. StreamAV-Bench outputs have resolution $1344\times768$.

\paragraph{StreamAV-Bench.}
Following the experimental setup of Vidu S2~\citep{zhang2026vidu}, we use all 160 Progressive and 160 Interactive scenarios of StreamAV-Bench~\citep{liu2026streamavbench}. The metrics cover instruction fulfillment (VIF, AIF), visual quality and aesthetics (VQ, VA), audio quality (AQ, PQ), audiovisual alignment and synchronization (AVAlign, AVSync), and subject and background consistency (SC, BC). All metrics are higher-is-better except AVSync, which measures temporal offset error in seconds. The 12-model comparison uses Progressive cases for VIF, AIF, SC, and BC, and both tracks for the remaining metrics.

\paragraph{Human evaluation.}
We invited experts with backgrounds in art or aesthetics to provide preference ratings for 70 pairs of image-to-video (I2V) sequences across five dimensions: consistency, instruction following, audiovisual synchronization, video quality, and motion quality. Good/Same/Bad (GSB) denote wins, ties, and losses relative to S2-Avatar at 3 NFE. Net preference is $100(G-B)/70$, retaining ties in the denominator. Usability requires Mean Opinion Scores (MOS) above 2 for both instruction following and synchronization.

\subsection{Results}
\label{sec:video_results}

\paragraph{Qualitative comparison.}
\figref{fig:qualitative_main} compares 3-NFE I2V outputs using the same reference images and timestamps. APEX retains both bags late in one sequence and moves the tube toward the other hand for the applying instruction in another, illustrating object persistence and a prompt-directed change in hand pose.
\input{figures/3_qualitative_main.tex}

\paragraph{Benchmark results.}
APEX ranks first on VA, VIF, and AIF and second on VQ among the 12 models with available results in \tabref{tab:streamav_overall}. In \tabref{tab:streamav_tracks}, VIF, AIF, and VA improve on both tracks, with Progressive VIF and AIF increasing by 0.025 and 0.067. VQ rises on Progressive ($+0.049$) but falls on Interactive ($-0.070$), yielding $-0.011$ overall. BC shows the opposite track-level pattern. AQ, AVAlign, and SC decrease on both tracks, pooled PQ is effectively unchanged, and pooled AVSync error rises from 1.319 to 1.341\,s. These results show instruction and aesthetic gains alongside visual and audiovisual trade-offs.
\input{tables/1_streamav_comparison.tex}
\input{tables/2_streamav_track_details.tex}

\paragraph{Human evaluation.}
\figref{fig:video_preference} and \tabref{tab:human_absolute} show the strongest preference for instruction following: 22 wins, 47 ties, and 1 loss yield $+30.0\%$ net preference, with MOS of 4.17 versus 3.99. Consistency and video quality each receive $+4.3\%$, corresponding to three net wins. Audiovisual synchronization is tied in 69 of 70 pairs, with one loss ($-1.4\%$) and MOS of 4.23 versus 4.26. Motion quality favors S2-Avatar ($-8.6\%$). Both models meet the usability criterion for 62 of 70 evaluated I2V sequences ($88.6\%$).
\input{figures/4_human_preferences.tex}
\input{tables/3_human_absolute.tex}

\subsection{Training and Sampling Comparisons}
\label{sec:video_additional}
Additional preference evaluations use separate rounds of 70 paired 180-second I2V sequences, each against S2-Avatar at 3 NFE with the same Refiner. \tabref{tab:training_comparisons} groups results by comparison round.

\paragraph{Training cost.}
In a separate measurement under matched training settings (\figref{fig:training_cost}), peak allocated memory decreases from 102.7 to 98.2\,GiB per GPU, saving 4.5\,GiB ($4.4\%$) relative to DMD. Mean iteration time decreases from 35.9 to 32.6\,s, a reduction of 3.3\,s ($9.2\%$), while retaining the frozen teacher.
\input{figures/5_training_cost.tex}
\input{figures/6_nfe2_human_preferences.tex}

\paragraph{Two-step generation.}
Against S2-Avatar at 3 NFE, APEX at 2 NFE receives $+27.1\%$ instruction preference (22 wins, 45 ties, 3 losses), while S2-Avatar at 2 NFE receives $+2.9\%$ in a separate paired round, as shown in \figref{fig:nfe2_preference}. APEX's other margins are negative: $-2.9\%$ for consistency and video quality, and $-7.1\%$ for synchronization and motion quality. \figref{fig:qualitative_nfe2} compares APEX at 2 NFE with S2-Avatar at 3 NFE, showing broadly similar character appearance with differences in pose and fine detail.
\input{figures/7_qualitative_nfe2.tex}

\paragraph{Timestep sampling.}
We vary the sampling of replay times $t_i$ over $(0.75,0.9375,0.99)$, leaving the scoring-time schedule for $t'$ unchanged. The default uses equal weights, whereas DF4 and DF10 use $(4,1,1)$ and $(10,1,1)$, raising the probability of drawing the lowest noise level from $1/3$ to $2/3$ and $5/6$. Sampled times are sorted along the replay segment before the differentiable pass. Default sampling has the highest instruction preference at both budgets in \tabref{tab:training_comparisons}, reaching $+27.1\%$ at 2 NFE and $+21.4\%$ at 3 NFE. Other dimensions vary across both inference budgets, with no consistent advantage from favoring lower-noise replay.

\paragraph{Optimizer configurations.}
\tabref{tab:training_comparisons} reports results for shared optimizer states at 3 NFE and separate generator/fake optimizer states at 2 NFE. Both configurations use AdamW with shared model parameters, achieving instruction-following net preferences of $+21.4\%$ and $+14.3\%$, respectively.
\input{tables/4_training_comparisons.tex}

%% file: figures/3_qualitative_main.tex
\begin{figure}[!t]
\centering
\includegraphics[width=\linewidth]{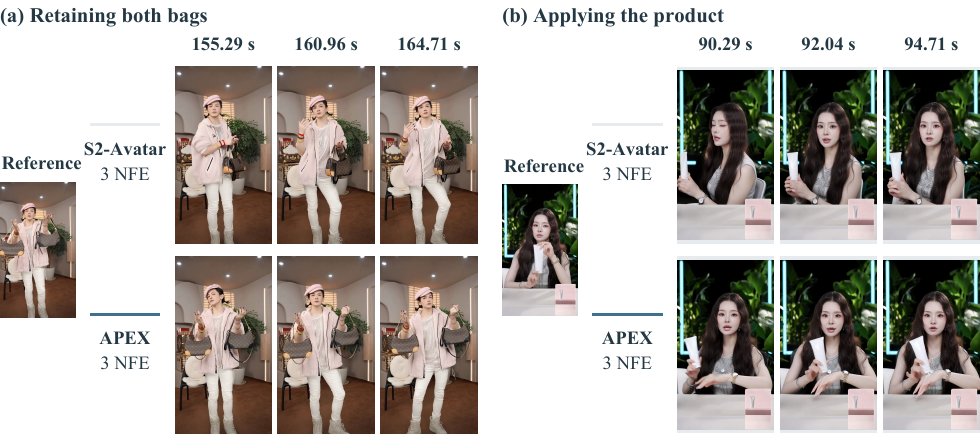}
\caption{\textbf{Paired examples at 3 backbone NFE.} (a) APEX retains both bags late in the sequence, while S2-Avatar shows one. (b) APEX brings the tube to the other hand for the applying instruction; S2-Avatar continues holding it at its side. Each pair uses the same reference and timestamps, with a common one-step $2\times$ Refiner.}
\label{fig:qualitative_main}
\end{figure}

%% file: tables/1_streamav_comparison.tex
\begin{table}[!ht]
\centering
\caption{\textbf{Comparison on ten StreamAV-Bench metrics.} VIF, AIF, SC, and BC use Progressive; the other metrics use both tracks. APEX and S2-Avatar use our evaluation results; baseline VIF is from StreamAV-Bench~\citep{liu2026streamavbench} and other baseline scores from Vidu S2~\citep{zhang2026vidu}. Values are rounded to three decimals. Bold and underlined values mark the best and second-best available results in each column; a dash marks an unreported value. AVSync uses seconds.}
\label{tab:streamav_overall}
\begingroup
\fontsize{8.6}{10.5}\selectfont
\renewcommand{\arraystretch}{1.12}
\apexfilledtable{11}{l*{10}{r}}{
\toprule
\rowcolor{apextablehead}Model & VA $\uparrow$ & VQ $\uparrow$ & PQ $\uparrow$ & AQ $\uparrow$ & AVAlign $\uparrow$ & AVSync $\downarrow$ & VIF $\uparrow$ & AIF $\uparrow$ & SC $\uparrow$ & BC $\uparrow$ \\
\midrule
SoulX-FlashTalk & 0.281 & 2.515 & \textbf{7.091} & \underline{3.006} & 0.121 & \textbf{0.648} & \multicolumn{1}{c}{--} & 2.499 & 0.946 & 0.964 \\
AptAvatar & 0.458 & 2.040 & \underline{7.075} & 2.582 & 0.099 & 1.284 & \multicolumn{1}{c}{--} & 2.448 & 0.860 & 0.923 \\
OmniForcing & 0.554 & 2.374 & 6.351 & 2.528 & 0.119 & 1.423 & 2.423 & 2.343 & 0.962 & 0.953 \\
Hallo-Live & 0.574 & \multicolumn{1}{c}{--} & 5.396 & \multicolumn{1}{c}{--} & 0.102 & 1.353 & \multicolumn{1}{c}{--} & \multicolumn{1}{c}{--} & \textbf{0.991} & \textbf{0.977} \\
Self-Forcing & 0.585 & 2.753 & 6.453 & 2.623 & 0.256 & 0.919 & 3.019 & \underline{2.810} & 0.981 & 0.969 \\
IAMFlow & 0.602 & 2.840 & 6.415 & 2.625 & \textbf{0.268} & 1.016 & 3.192 & 2.789 & 0.986 & 0.973 \\
AvatarForcing & \underline{0.639} & \multicolumn{1}{c}{--} & 7.070 & \multicolumn{1}{c}{--} & 0.089 & 1.041 & \multicolumn{1}{c}{--} & \multicolumn{1}{c}{--} & \underline{0.987} & \underline{0.976} \\
PixVerse R1 & 0.529 & 2.428 & 6.348 & 2.810 & 0.234 & \underline{0.855} & 2.976 & 2.511 & 0.907 & 0.930 \\
Odyssey-2 & 0.531 & 2.820 & 6.426 & 2.659 & \underline{0.260} & 0.946 & 3.106 & 2.778 & 0.976 & 0.966 \\
HappyOyster & 0.529 & 2.785 & 6.817 & \textbf{3.157} & 0.206 & 1.044 & 3.565 & 2.692 & 0.904 & 0.934 \\
\midrule
Vidu S2-Avatar & 0.633 & \textbf{3.267} & 6.386 & 2.788 & 0.103 & 1.319 & \underline{3.831} & 2.768 & 0.979 & 0.972 \\
\rowcolor{apextableours}\textbf{APEX (ours)} & \textbf{0.642} & \underline{3.256} & 6.386 & 2.725 & 0.101 & 1.341 & \textbf{3.856} & \textbf{2.835} & 0.978 & 0.972 \\

\bottomrule}
\endgroup
\end{table}

%% file: tables/2_streamav_track_details.tex
\begin{table}[!ht]
\centering
\caption{\textbf{Per-track and pooled StreamAV-Bench results.} Progressive and Interactive each contain 160 cases; Overall pools all 320 cases for every metric. Values retain source precision, and $\Delta$ is APEX minus S2-Avatar. AVSync is measured in seconds and is lower-is-better; all other reported metrics are higher-is-better.}
\label{tab:streamav_tracks}
\begingroup
\fontsize{8.3}{10.2}\selectfont
\renewcommand{\arraystretch}{1.12}
\apexfilledtable{10}{l*{9}{r}}{
\toprule
\rowcolor{apextablehead}Metric & \multicolumn{3}{c}{Overall ($n=320$)} & \multicolumn{3}{c}{Progressive ($n=160$)} & \multicolumn{3}{c}{Interactive ($n=160$)} \\
\cmidrule(lr){2-4}\cmidrule(lr){5-7}\cmidrule(lr){8-10}
 & APEX & S2 & $\Delta$ & APEX & S2 & $\Delta$ & APEX & S2 & $\Delta$ \\
\midrule
VQ $\uparrow$ & 3.256 & 3.267 & $-0.011$ & 3.180 & 3.131 & $+0.049$ & 3.333 & 3.403 & $-0.070$ \\
AQ $\uparrow$ & 2.725 & 2.788 & $-0.063$ & 2.706 & 2.809 & $-0.103$ & 2.744 & 2.766 & $-0.022$ \\
VIF $\uparrow$ & 3.891 & 3.834 & $+0.057$ & 3.856 & 3.831 & $+0.025$ & 3.925 & 3.836 & $+0.089$ \\
AIF $\uparrow$ & 2.610 & 2.570 & $+0.040$ & 2.835 & 2.768 & $+0.067$ & 2.385 & 2.372 & $+0.013$ \\
\midrule
VA $\uparrow$ & 0.641898 & 0.632606 & $+0.009292$ & 0.645374 & 0.636402 & $+0.008972$ & 0.638422 & 0.628810 & $+0.009612$ \\
PQ $\uparrow$ & 6.385856 & 6.385855 & $+0.000001$ & 6.380547 & 6.380545 & $+0.000002$ & 6.391166 & 6.391166 & $+0.000000$ \\
AVAlign $\uparrow$ & 0.101355 & 0.103383 & $-0.002028$ & 0.100186 & 0.102011 & $-0.001825$ & 0.102524 & 0.104754 & $-0.002230$ \\
AVSync $\downarrow$ & 1.340521 & 1.318854 & $+0.021667$ & 1.357500 & 1.323646 & $+0.033854$ & 1.323542 & 1.314062 & $+0.009480$ \\
SC $\uparrow$ & 0.979032 & 0.979483 & $-0.000451$ & 0.978310 & 0.978624 & $-0.000314$ & 0.979754 & 0.980342 & $-0.000588$ \\
BC $\uparrow$ & 0.972869 & 0.972753 & $+0.000116$ & 0.971695 & 0.972411 & $-0.000716$ & 0.974043 & 0.973096 & $+0.000947$ \\

\bottomrule}
\endgroup
\end{table}

%% file: figures/4_human_preferences.tex
\begin{figure}[!htb]
\centering
\includegraphics[width=\linewidth]{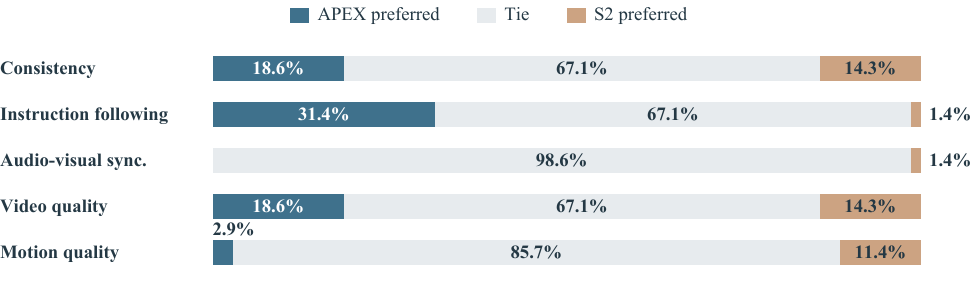}
\caption{\textbf{Paired human preferences at 3 backbone NFE.} APEX versus S2-Avatar on 70 I2V sequences of 180 seconds, with a common one-step $2\times$ Refiner. Good, Same, and Bad denote APEX wins, ties, and losses.}
\label{fig:video_preference}
\end{figure}

%% file: tables/3_human_absolute.tex
\begin{table}[!ht]
\centering
\caption{\textbf{Absolute human ratings and usability.} Usable sequences have both MOS ratings above 2.}
\label{tab:human_absolute}
\begingroup
\fontsize{9}{11}\selectfont
\renewcommand{\arraystretch}{1.12}
\begin{tabular}{lrrrr}
\toprule
\rowcolor{apextablehead}Model & Instruction MOS & Synchronization MOS & Usable sequences & Usability \\
\midrule
APEX & \textbf{4.17} & 4.23 & 62/70 & 88.6\% \\
S2-Avatar & 3.99 & \textbf{4.26} & 62/70 & 88.6\% \\
\bottomrule
\end{tabular}
\endgroup
\end{table}

%% file: figures/5_training_cost.tex
\begin{figure}[!htb]
\centering
\includegraphics[width=\linewidth]{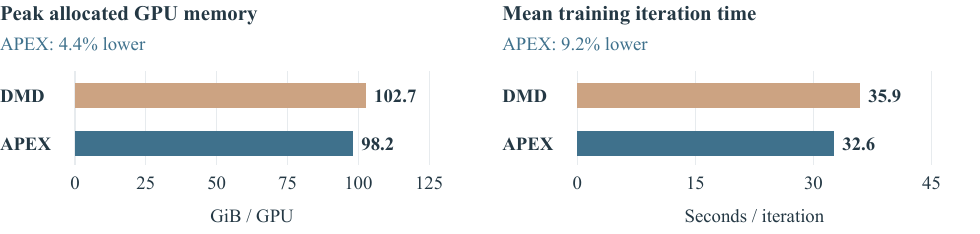}
\caption{\textbf{Training cost of DMD and APEX.} Peak allocated memory per GPU (GiB) and mean iteration time (seconds) under matched training settings. Percentage reductions are measured relative to DMD.}
\label{fig:training_cost}
\end{figure}

%% file: figures/6_nfe2_human_preferences.tex
\begin{figure}[!htb]
\centering
\includegraphics[width=\linewidth]{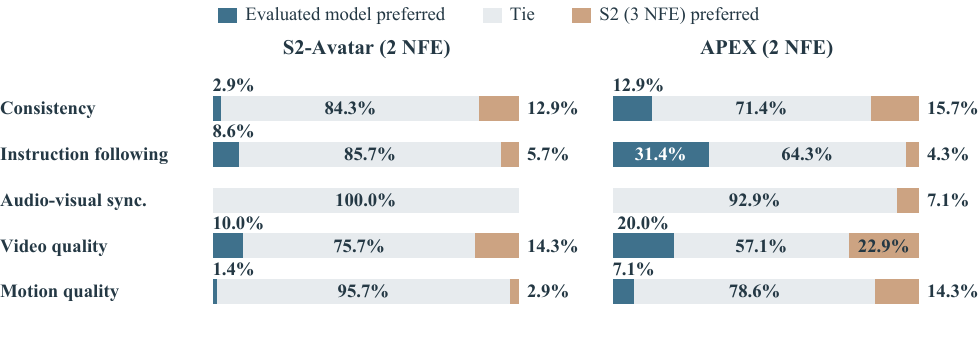}
\caption{\textbf{Human preferences at 2 backbone NFE.} Each model is evaluated against S2-Avatar at 3 NFE in a separate 70-case paired comparison. All configurations use the same one-step $2\times$ Refiner.}
\label{fig:nfe2_preference}
\end{figure}

%% file: figures/7_qualitative_nfe2.tex
\begin{figure}[!htb]
\centering
\includegraphics[width=\linewidth]{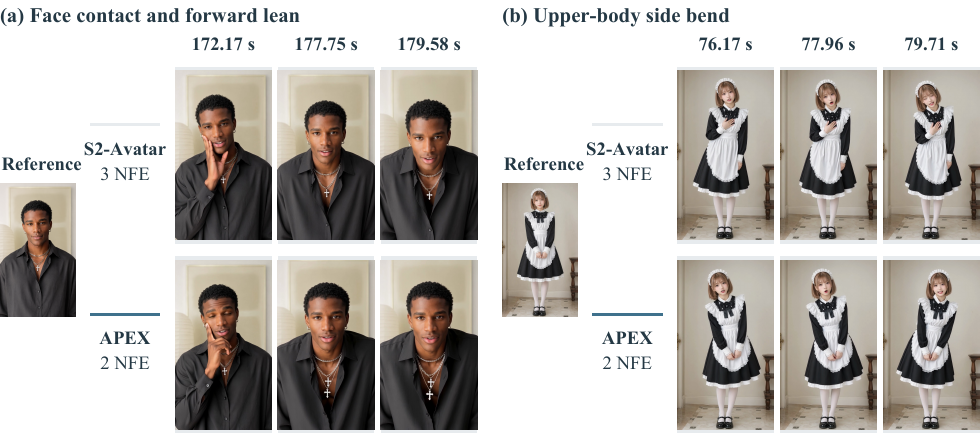}
\caption{\textbf{APEX at 2 backbone NFE versus S2-Avatar at 3.} Selected face-contact, forward-leaning, and side-bending frames have similar overall appearance, with differences in jewelry, hand placement, and body pose. Reference images, frame timestamps, and the one-step $2\times$ spatial Refiner are matched across methods.}
\label{fig:qualitative_nfe2}
\end{figure}

%% file: tables/4_training_comparisons.tex
\begin{table}[!htbp]
\centering
\caption{\textbf{Sampling and optimizer configurations.} Net preference (\%) over S2-Avatar at 3 NFE; each block is a separate 70-case round. Sampling variants share optimizer states; optimizer rows use AdamW and also differ in NFE.}
\label{tab:training_comparisons}
\begingroup
\fontsize{9}{11}\selectfont
\renewcommand{\arraystretch}{1.12}
\apexfilledtable{7}{lcrrrrr}{
\toprule
\rowcolor{apextablehead}Configuration & NFE & Consistency & Instruction & Sync. & Video & Motion \\
\midrule
Default & 2 & $-2.9$ & $+27.1$ & $-7.1$ & $-2.9$ & $-7.1$ \\
DF10 & 2 & $-18.6$ & $+21.4$ & $-5.7$ & $-8.6$ & $-4.3$ \\
DF4 & 2 & $-8.6$ & $+17.1$ & $-8.6$ & $-11.4$ & $-8.6$ \\
\midrule
Default & 3 & $-5.7$ & $+21.4$ & $-7.1$ & $-5.7$ & $+2.9$ \\
DF10 & 3 & $-11.4$ & $+11.4$ & $-7.1$ & $-11.4$ & $+1.4$ \\
DF4 & 3 & $-14.3$ & $+10.0$ & $-10.0$ & $-7.1$ & $+1.4$ \\
\midrule
Shared states & 3 & $+5.7$ & $+21.4$ & $-14.3$ & $+2.9$ & $-7.1$ \\
Separate states & 2 & $-8.6$ & $+14.3$ & $-15.7$ & $-10.0$ & $-7.1$ \\
\bottomrule
}
\endgroup
\end{table}

%% file: sections/5_conclusion.tex
\begin{samepage}
\section{Conclusion}
APEX integrates condition-shifted fake estimation into S2's Self-Replay Forcing, sharing generator and fake parameters while retaining the frozen teacher. It improves instruction fulfillment and visual aesthetics, with positive human preference for instruction following at both 3 and 2 backbone NFE. Matched measurements show lower peak allocated GPU memory and shorter mean iteration time. Motion quality and audiovisual synchronization remain trade-offs, while APEX and S2-Avatar achieve the same usability rate.
\par
\end{samepage}
\clearpage

%% file: sections/6_related_work.tex
\section{Related Work}

\subsection{Consistency and Trajectory-Based Generation}
\label{sec:rw_path}
Diffusion~\citep{ho2020denoising,song2020score,karras2022edm} and flow matching~\citep{lipman2022flow} generate samples by integrating a learned velocity field from noise to data.
Embedded Representation Warmup~\citep{liu2025efficient} improves training efficiency by aligning early model layers with a pretrained encoder before full generative training.
Consistency and trajectory-based methods train the model to predict over a time interval, allowing generation with fewer network evaluations.
Consistency Models (CMs)~\citep{song2023consistency,lu2024simplifying} learn to map points on the same trajectory to a common endpoint.
Score-regularized continuous-time consistency (rCM) combines consistency training with score-based distribution matching for large-scale diffusion distillation~\citep{zheng2025rcm}.
MeanFlow~\citep{geng2025mean} predicts the average velocity over an interval, while Transition Models (TiM)~\citep{wang2025transition} learn the transition between its endpoints.
DuMo~\citep{sun2026duality} jointly predicts local velocity and a flow map with two heads on a shared backbone.
UCGM~\citep{sun2025unified} interpolates between generative training objectives.

\subsection{Distribution Matching and Adversarial Training}
\label{sec:rw_endpoint}
Few-step generation also benefits from direct feedback on the generated distribution.

External correction signals can come from auxiliary score models, as in distribution matching~\citep{yin2024one,yin2024improved}, or from discriminators used in adversarial refiners~\citep{kim2023refining,sauer2024fast,zheng2025direct}.
Additional networks increase memory use and update cost. Some early distillation methods also use precomputed teacher-student pairs for regularization, while DMD2 removes that paired-data requirement~\citep{yin2024improved}.
Direct Discriminative Optimization (DDO)~\citep{zheng2025direct} expresses the GAN discriminator as a likelihood ratio constructed from the target and a fixed reference model.

TwinFlow~\citep{cheng2026twinflow} learns real and fake trajectories within a shared model and uses their velocity discrepancy to construct an internal correction signal.
LLaDA-Image Turbo~\citep{chen2026llada} adapts TwinFlow-based distillation to support image generation in two to four sampling steps.

\APEX~\citep{liu2026self} follows this line and separates the branches through condition shifting. The shifted condition selects the fake estimator for samples generated with the original prompt. In \secref{subsec:gan_connection}, our video formulation shares generator and fake parameters while retaining S2's frozen reference teacher.

Intrinsic Distribution Matching (IDM)~\citep{liu2026intrinsic} adds feature mean and standard deviation matching to DMD using the generator's exponential moving average (EMA) copy.

Few-step score identity distillation (SiD)~\citep{zhou2025few} also studies how conditioning affects distribution matching. Its Zero-CFG strategy removes text conditioning from the fake-score network, while Anti-CFG applies negative classifier-free guidance by combining conditional and unconditional predictions. These strategies modify the fake-score estimate. \APEX fits the fake velocity under shifted text.

Fake-Score-network-Free DMD (FSF-DMD)~\citep{kim2026distribution} replaces the auxiliary fake-score network with a generator-induced endpoint pseudo-velocity. \APEX instead learns the fake velocity through flow matching on generated samples under a shifted condition.
TBSM~\citep{sun2026three} constructs regression targets from attraction to real samples and repulsion from generated samples under an energy-distance objective; its correction is defined by sample interactions rather than a real/fake velocity difference.

\subsection{Causal Streaming Video}
Autoregressive video diffusion has been studied through asynchronous generation~\citep{sun2025ar}, progressive prediction~\citep{xie2025progressive}, and causal generation with cache sharing~\citep{gao2025ca2vdmefficientautoregressivevideo}.
MAGI-1 studies autoregressive generation at scale~\citep{teng2025magi}, while SkyReels-V2 targets extended video generation~\citep{chen2025skyreels}.
FIFO-Diffusion~\citep{kim2024fifo} and StreamingT2V~\citep{henschel2025streamingt2v} also investigate extending video generation beyond a short clip.
Later predictions depend on generated history.

CausVid distills bidirectional video diffusion into an autoregressive model~\citep{yin2025slow}; diagonal distillation provides another streaming autoregressive formulation~\citep{liu2026streaming}.
Causal Forcing and Causal Forcing++ study few-step autoregressive distillation for real-time interaction~\citep{zhu2026causal,zhao2026causal}.
Causal-rCM unifies Teacher Forcing and Self Forcing in a streaming distillation recipe~\citep{zheng2026causal}.
LPM studies video-based character performance~\citep{zeng2026lpm}, providing a setting for distribution matching in autoregressive models that animate characters.

S2-Avatar~\citep{zhang2026vidu} combines bidirectional reference-conditioned training, causal adaptation with Teacher and Diffusion Forcing, and Self-Replay Forcing.
Diffusion Forcing~\citep{DF} exposes the model to noisy historical states, while Self Forcing~\citep{huang2026self} trains with autoregressive histories generated by the current model.
S2's replay procedure retains cross-block gradients inside a sampled trajectory segment without retaining the full rollout graph.
APEX uses this video setting to learn the fake estimator within the generator while retaining S2's causal replay procedure and reference teacher.
FlowCPO~\citep{han2026flowcpo} studies offline preference alignment using fixed preferred and dispreferred samples.
Preference optimization and latent refinement are separate from the APEX correction.

\subsection{Historical Context and Long Video Generation}
History-Guided Video Diffusion studies conditioning on past video~\citep{song2025history}, while Grounded Forcing separates persistent semantics from recent dynamics~\citep{chen2026grounded}.
LongLive~\citep{yang2025longlive} and Rolling Forcing~\citep{liu2025rolling} target real-time long video generation; Infinity-RoPE investigates action control under autoregressive self-rollout~\citep{yesiltepe2026infinity}.
UltraViCo~\citep{zhao2026ultravico} addresses length extrapolation by suppressing attention dispersion beyond the training window, providing a complementary approach to extending video duration.
These works motivate evaluating instruction response and motion quality alongside visual appearance as a stream grows longer.

Memory design provides a complementary approach.
SWIFT uses prompt-adaptive memory~\citep{tan2026swift}, IAMFlow studies identity-aware memory~\citep{liu2026iamflow}, and FramePack compresses the input-frame context for next-frame prediction~\citep{zhang2025packing}.
Deep Forcing uses sink states and compression~\citep{yi2025deep}; Rolling Sink studies the gap between limited training horizons and open-ended testing~\citep{li2026rolling}.
MemRoPE uses evolving memory tokens for extended generation~\citep{kim2026memrope}.
Attention sinks were also studied for streaming language models~\citep{xiao2024efficient}; S2 adopts sink blocks within its video context design~\citep{vidus1,zhang2026vidu}.
APEX retains this context design for its shared model.
Quant VideoGen~\citep{xi2026quant} reduces autoregressive history-storage costs through two-bit KV-cache quantization, using semantic-aware smoothing and progressive residual quantization.
ViewRope~\citep{xiang2026geometry} instead injects camera-ray geometry into rotary position embeddings and selects relevant historical frames to improve spatial consistency in camera-controlled video world models.
These methods address history storage and retrieval, complementing APEX's focus on the training cost of distribution estimation.

For streaming video understanding, ChronoReplay~\citep{hu2026chronoreplay} organizes heterogeneous spatial supervision into chronological episodes, while \emph{Supervising the Visible State}~\citep{hu2026supervising} separates supervision of temporal state from adaptive selection of visual evidence.
These studies provide complementary context for the role of temporal supervision and historical evidence in streaming video understanding.

\subsection{Audio-Driven and Interactive Video}
Audio-driven character generation includes portrait animation with Hallo3~\citep{cui2025hallo3}, body animation with OmniAvatar~\citep{gan2025omniavatar}, and cinematic video with Wan-S2V~\citep{gao2025wan}.
StableAvatar studies extended audio-driven generation~\citep{tu2025stableavatar}.
Live Avatar~\citep{huang2025live}, SoulX-FlashTalk~\citep{shen2025soulxflashtalk}, and AptAvatar~\citep{zhang2026aptavatar} address real-time or long-form avatar generation.
AvatarForcing specifically studies one-step streaming talking avatars~\citep{cui2026avatarforcing}.
Joint audio and video generation is studied by OmniForcing~\citep{su2026omniforcing}, Hallo-Live~\citep{li2026hallolive}, and Omni-LiveAvatar~\citep{zhu2026omniliveavatar}.
TurboT2VA~\citep{yang2026turbot2va} extends consistency distillation to joint video-audio generation with modality-specific normalization and a progressive curriculum that introduces distribution matching after consistency training.
This connects few-step distillation to audiovisual synchronization, although its joint-generation setting differs from the causal replay setting studied here.
Comparisons must distinguish joint audio generation from video generation conditioned on fixed external audio.

S2's evaluation also includes external systems such as PixVerse R1~\citep{pixverse2026r1}, HappyOyster~\citep{alibaba2026happyoyster}, Odyssey-2~\citep{odyssey2025odyssey2}, and HeyGen~\citep{heygen2026avatar}.
Their inclusion in S2 provides evaluation context; the APEX evidence here consists of the internal I2V comparison and the completed StreamAV-Bench run in \secref{sec:exp}.
Comparisons must match references, audio, duration, and resolution.

\subsection{Efficient Video Inference}
Sampling steps are one component of generation cost.
TurboDiffusion accelerates video diffusion~\citep{zhang2025turbodiffusion}, and TurboServe addresses efficient streaming service~\citep{jiang2026turboserve}.
Pyramidal Flow Matching studies efficient video generation across scales~\citep{jin2025pyramidal}; the present experiments account separately for the low-resolution backbone and one-step high-resolution spatial Refiner.

Efficient attention methods can be organized into hardware-efficient, sparse, compact, and linear approaches~\citep{zhang2025efficientattention}.
Sparse VideoGen~\citep{xi2025sparse} exploits spatial and temporal attention-head patterns, while Sparse VideoGen2~\citep{yang2025sparsevideogen2} uses semantic-aware token permutation to identify and group important attention blocks.
SVG-EAR~\citep{zhou2026svgear} complements sparse computation with parameter-free centroid compensation and error-aware routing for skipped blocks.

S2's attention implementation draws on several families of efficient operators.
SageAttention and SageAttention2 use low-bit attention~\citep{zhang2025sageattention,zhang2024sageattention2}; SageAttention2++ improves its implementation~\citep{zhang2025sageattention2++}, and SageAttention3 explores FP4 inference and low-bit training~\citep{zhang2025sageattention3}.
SageBwd further studies trainable low-bit attention~\citep{zhang2026sagebwd}.
SpargeAttention studies sparse attention without additional training~\citep{zhang2025spargeattention}, while SpargeAttention2 uses trainable masking and distillation~\citep{zhang2026spargeattention2}.
SLA and SLA2 combine sparse and linear attention. SLA2 incorporates learnable routing together with quantization-aware training~\citep{zhang2025sla,zhang2026sla2}.

At the linear-layer level, 6Bit-Diffusion~\citep{su2026sixbit} dynamically allocates NVFP4 and INT8 precision according to quantization sensitivity and reuses temporally stable block residuals.
Speculative Decoding for Autoregressive Video Generation~\citep{hu2026speculative} uses a smaller model to draft video blocks and a quality router to decide which blocks require regeneration by the larger target model.
Both affect inference computation, a separate consideration from the number of networks maintained during distillation.

S2 also draws on block-level low-bit computation~\citep{zhang2025accurate} and Ulysses-style parallelism~\citep{jacobs2023deepspeed}.
These operators, precision choices, and parallel schedules can change latency at the same NFE. The training-cost comparison in \figref{fig:training_cost} measures APEX and DMD under matched settings.

\subsection{Streaming Video Editing}
StreamDiffusionV2 studies dynamic interactive video generation~\citep{feng2025streamdiffusionv2}.
LiveEdit~\citep{wang2026liveedit}, JoyAI-Video-Edit~\citep{xiao2026joyaivideoedit}, and SANA-Streaming~\citep{zhao2026sanastreaming} address streaming video editing.
These methods use an incoming video's motion and structure, which differs from I2V conditioning.

\paragraph{Interactive streaming evaluation.}
StreamEval~\citep{team2026streameval} evaluates timed interactions across avatar dialogue, avatar performance, world generation, instruction editing, and reference editing, measuring task execution, state consistency, visual fidelity, and audiovisual coordination.
Our reported benchmark results use StreamAV-Bench~\citep{liu2026streamavbench} and do not include StreamEval measurements.

\paragraph{Visual text evaluation.}
VTR-Bench~\citep{huang2026vtr} evaluates visual text fidelity separately from scene and motion requirements in video.
UltraText Bench~\citep{liu2026ultratext} evaluates dense bilingual text in individual images, providing complementary background on visual text rendering.
Neither is evaluated here.

%% file: sections/7_theoretical_analysis.tex
\section{Theoretical Analysis and Proofs}
\label{app:theory}

We derive the OT score-velocity identity and the frozen-target gradient in \secref{sec:apex}. The endpoint formulation in \appref{app:legacy_endpoint} compares clean-latent and velocity losses algebraically. The score identities fix conditioning; the replay gradient differentiates connected predictions with the target fixed. These results do not establish long-horizon stability or an exact KL gradient for the joint video trajectory.

\subsection{Setup}
\label{app:notation}

We use bold lowercase letters for vectors like $\xx, \zz, \vv$ and bold uppercase letters for matrices and operators like $\mmF$.
The identity matrix is denoted by $\mI$, and $\mathbf{0}$ represents the zero vector.

For $\xx,\zz\in\R^d$, let $p_r(\xx)$ be the data distribution and $p_z(\zz)=\cN(\mathbf{0},\mI)$ the standard Gaussian prior.

For conditional generation, $\cc$ denotes the projected text representation, and $p_r(\xx\mid\cc)$ is the conditional data distribution. Additional video conditions $\hh$ are held fixed and suppressed, as is $\cc$ when unambiguous.
Throughout this appendix, we work with the Optimal Transport (OT) interpolation path defined by:
\begin{equation}
\xx_t = t\zz + (1-t)\xx, \quad t \in [0,1],
\label{eq:ot_path_app}
\end{equation}
This satisfies the boundary conditions: $\xx_0 = \xx$ (pure data) and $\xx_1 = \zz$ (pure noise).
All score identities and time weights below are used for $t\in(0,1)$.
The per-sample FM target is constant along this path:
\begin{equation}
\uu_r(\xx,\zz) := \zz-\xx.
\end{equation}

The population real-data velocity is the conditional mean of this target:
\begin{equation}
\vv_r^*(\xx_t,t) := \mathbb E[\uu_r(\xx,\zz)\mid\xx_t,t].
\end{equation}

The score function is $\ss_{r,t}(\xx_t) := \nabla_{\xx_t} \log p_{r,t}(\xx_t)$, where $p_{r,t}(\xx_t)$ is the marginal density of $\xx_t$ at time $t$.

A star marks a population conditional mean, distinct from a network estimate.

We write $\mmF_{\theta}$ for the velocity estimate $\mmF_{\theta}(\xx_t,t,\cc)\in\R^d$ when its inputs are clear.

The operator $\sg(\cdot)$ denotes stop-gradient. For $\theta\in\R^P$ and a vector output $\mathbf q\in\R^d$, $J_\theta\mathbf q:=\partial\mathbf q/\partial\theta\in\R^{d\times P}$, so the chain rule uses its transpose to map output derivatives to parameter gradients.

For video, branch $r$ is the guided frozen-teacher reference and branch $\theta$ is the shared fake estimate. Teacher parameters are $\phi$. Hats mark network estimates; unadorned scores are exact.

Throughout this paper, $\mmF_\theta(\yy,t,\cc)$ denotes a velocity prediction with no destination-time argument. The one-step map $\mmG_\theta(\zz,\cc)=\zz-\mmF_\theta(\zz,1,\cc)$ and S2's clean-latent conversion use this same parameterization.

We define the endpoint predictor that maps a velocity estimate to its implied clean sample:
\begin{equation}
\mf^{\xx}(\mmF, \xx_t, t) := \xx_t - t \cdot \mmF.
\label{eq:endpoint_pred_app}
\end{equation}
This is motivated by the OT path: if $\xx_t = t\zz + (1-t)\xx$ and $\mmF \approx \zz - \xx$, then $\mf^{\xx}(\mmF, \xx_t, t) \approx \xx$.

\subsection{Score-Velocity Duality under OT Path}
\label{app:score_velocity}

The OT path gives a score-velocity identity for both real and generated endpoint distributions.

\begin{proposition}[Score-Velocity Duality]
\label{prop:score_velocity}

Let $\xx_t=t\zz+(1-t)\xx$ for $t\in(0,1)$, with independent draws $\zz\sim\cN(\mathbf{0},\mI)$ and $\xx\sim p_r$.
Write $p_{r,t}$ for the density of $\xx_t$. The optimal least-squares velocity is
\begin{equation}
\vv_r^*(\xx_t,t) := \mathbb E[\uu_r(\xx,\zz)\mid\xx_t,t].
\label{eq:vstar_def}
\end{equation}

Then the score function $\ss_{r,t}(\xx_t) := \nabla_{\xx_t}\log p_{r,t}(\xx_t)$ along the OT path satisfies
\begin{equation}
\ss_{r,t}(\xx_t) = -\frac{\xx_t + (1-t)\vv_r^*(\xx_t,t)}{t}.
\label{eq:score_velocity_relation}
\end{equation}
\end{proposition}

\begin{proof}
\textbf{Step 1: Rewrite as an additive Gaussian observation model.}
Define $\xx':=(1-t)\xx$. Then the OT path becomes an additive Gaussian observation of the scaled endpoint:
\[
\xx_t=\xx' + t\zz,\qquad \zz\sim\cN(\mathbf{0},\mI).
\]
Conditioned on $\xx'$, the likelihood is $\xx_t\mid \xx' \sim \cN(\xx',t^2\mI)$, since $\xx_t - \xx' = t\zz$ and $\zz \sim \cN(\mathbf{0}, \mI)$ implies $t\zz \sim \cN(\mathbf{0}, t^2\mI)$, with the noise covariance determined by the sampled diffusion time.

\textbf{Step 2: Apply Tweedie's formula to recover the posterior mean.}
At fixed $t$, Tweedie's formula~\citep{efron2011tweedie} recovers the posterior mean of the scaled endpoint $\xx'=(1-t)\xx$ from the marginal score of $\xx_t=\xx'+t\zz$:
\begin{equation}
\mathbb E[\xx'\mid \xx_t] = \xx_t + t^2\,\nabla_{\xx_t}\log p_{r,t}(\xx_t) = \xx_t + t^2\,\ss_{r,t}(\xx_t).
\label{eq:tweedie_xprime}
\end{equation}
Since $\xx'=(1-t)\xx$, we recover the conditional expectation of the original clean endpoint $\xx$:
\begin{align}
\mathbb E[\xx\mid \xx_t]
&= \frac{1}{1-t}\mathbb E[\xx'\mid \xx_t] \nonumber\\
&= \frac{1}{1-t}\left[\xx_t + t^2\,\ss_{r,t}(\xx_t)\right] \nonumber\\
&= \frac{\xx_t + t^2\,\ss_{r,t}(\xx_t)}{1-t}.
\label{eq:cond_x}
\end{align}

\textbf{Step 3: Express the conditional mean of $\zz$.}
From the OT path $\xx_t=t\zz+(1-t)\xx$, solving for the noise gives:
\[
\zz=\frac{\xx_t-(1-t)\xx}{t}.
\]
Taking expectations conditioned on $\xx_t$ and substituting the mean endpoint gives:
\begin{align}
\mathbb E[\zz\mid \xx_t]
&= \mathbb E\!\left[\frac{\xx_t-(1-t)\xx}{t}\,\middle|\,\xx_t\right] \nonumber\\
&= \frac{1}{t}\Bigl[\xx_t - (1-t)\mathbb E[\xx\mid \xx_t]\Bigr] \nonumber\\
&= \frac{1}{t}\Bigl[\xx_t - (1-t)\cdot \frac{\xx_t + t^2\,\ss_{r,t}(\xx_t)}{1-t}\Bigr] \quad \text{substituting \equref{eq:cond_x}} \nonumber\\
&= \frac{1}{t}\Bigl[\xx_t - \bigl(\xx_t + t^2\,\ss_{r,t}(\xx_t)\bigr)\Bigr] \quad \text{(simplifying the fraction)} \nonumber\\
&= \frac{1}{t}\bigl[-t^2\,\ss_{r,t}(\xx_t)\bigr] \nonumber\\
&= -t\,\ss_{r,t}(\xx_t).
\label{eq:cond_z_clean}
\end{align}

\textbf{Step 4: Form the optimal velocity and rearrange.}
By definition, the (least-squares) optimal velocity field along the OT path is the conditional expectation of the target velocity $\zz - \xx$:
\[
\vv_r^*(\xx_t,t):=\mathbb E[\uu_r(\xx,\zz)\mid\xx_t,t]=\mathbb E[\zz\mid \xx_t]-\mathbb E[\xx\mid \xx_t].
\]
Substituting \equref{eq:cond_x} and \equref{eq:cond_z_clean} into the conditional-mean velocity definition gives
\begin{align}
\vv_r^*(\xx_t,t)
&= -t\,\ss_{r,t}(\xx_t) - \frac{\xx_t + t^2\,\ss_{r,t}(\xx_t)}{1-t} \nonumber\\
&= \frac{-t(1-t)\,\ss_{r,t}(\xx_t) - (\xx_t + t^2\,\ss_{r,t}(\xx_t))}{1-t} \quad \text{(common denominator)} \nonumber\\
&= \frac{-t\,\ss_{r,t}(\xx_t) + t^2\,\ss_{r,t}(\xx_t) - \xx_t - t^2\,\ss_{r,t}(\xx_t)}{1-t} \nonumber\\
&= \frac{-t\,\ss_{r,t}(\xx_t) - \xx_t}{1-t} \nonumber\\
&= -\frac{\xx_t + t\,\ss_{r,t}(\xx_t)}{1-t}.
\label{eq:vstar_in_terms_of_score}
\end{align}

\textbf{Step 5: Rearrange to obtain the score-velocity duality.}
Multiplying both sides of \equref{eq:vstar_in_terms_of_score} by $(1-t)$ gives an equivalent relation between the velocity and marginal score along the OT path:
\[
(1-t)\vv_r^*(\xx_t,t) = -\xx_t - t\,\ss_{r,t}(\xx_t).
\]
Moving both the sample and velocity terms to the left-hand side of the equality gives:
\[
\xx_t + (1-t)\vv_r^*(\xx_t,t) = -t\,\ss_{r,t}(\xx_t),
\]
which, upon dividing both sides by $-t$, gives the stated score-velocity identity in \equref{eq:score_velocity_relation}:
\[
\ss_{r,t}(\xx_t) = -\frac{\xx_t + (1-t)\vv_r^*(\xx_t,t)}{t}.
\]
\end{proof}

\begin{corollary}[Velocity Difference as Score Difference]
\label{cor:velocity_score}

For $i\in\{1,2\}$, let $p_{i,t}$ be the marginal of $\xx_t=(1-t)\xx+t\zz$ under endpoint law $i$ and independent Gaussian noise. Define $\vv_i^*(\yy,t):=\mathbb E_i[\zz-\xx\mid\xx_t=\yy,t]$ and $\ss_{i,t}(\yy):=\nabla_{\yy}\log p_{i,t}(\yy)$. At the same noisy state and the same diffusion time,
\begin{equation}
\vv_1^*(\xx_t,t) - \vv_2^*(\xx_t,t) = -\frac{t}{1-t}\left[\ss_{1,t}(\xx_t) - \ss_{2,t}(\xx_t)\right].
\label{eq:velocity_score_diff}
\end{equation}
\end{corollary}

\begin{proof}
Applying \propref{prop:score_velocity} to both fields at the same state and time gives:
\begin{align}
\ss_{1,t}(\xx_t) &= -\frac{\xx_t + (1-t)\vv_1^*(\xx_t,t)}{t}, \label{eq:score1} \\
\ss_{2,t}(\xx_t) &= -\frac{\xx_t + (1-t)\vv_2^*(\xx_t,t)}{t}. \label{eq:score2}
\end{align}
Subtracting \equref{eq:score2} from \equref{eq:score1} cancels the common state term at the same time:
\begin{align}
\ss_{1,t}(\xx_t) - \ss_{2,t}(\xx_t)
&= -\frac{1}{t}\left[(\xx_t + (1-t)\vv_1^*) - (\xx_t + (1-t)\vv_2^*)\right] \nonumber \\
&= -\frac{1-t}{t}\left[\vv_1^*(\xx_t,t) - \vv_2^*(\xx_t,t)\right].
\end{align}
Rearranging yields \equref{eq:velocity_score_diff} for the difference between the two velocity fields.
\end{proof}

\subsection{Frozen-Target Correction}
\label{app:apex_velocity_correction}

Suppressing block indices, let $\widehat\xx_{0,r}$ be the guided prediction of the frozen teacher and $\widehat\xx_{0,\theta}=\xx_{\theta,t'}^{\sg}-t'\widehat\vv_{\theta,t'}$ the fake clean-latent prediction. The DMD correction used in \equref{eq:video_correction} is
\begin{equation}
 w:=\operatorname{mean}\!\left|\xx_\theta-\widehat\xx_{0,r}\right|,\qquad
 \widetilde\dd_{t'}:=\frac{\widehat\xx_{0,r}-\widehat\xx_{0,\theta}}{w}
 =\frac{t'}{w}\left(\widehat\vv_{\theta,t'}-\widehat\vv_{r,t'}\right).
\label{eq:video_normalized_correction}
\end{equation}
The mean uses the inherited video mask and is computed per video across the full scored sequence, including any detached rollout prefix. It is not recomputed separately for each block. Both estimator predictions and the normalization $w$ stay fixed when differentiating the generator loss with respect to $\theta$.

For the derivative, write $\xx_\theta$ and $\widetilde\dd_{t'}$ as the masked vectors of $d$ coordinates used by the MSE, and let $J:=\partial\xx_\theta/\partial\theta$. Differentiating only the prediction outside $\sg$ in \equref{eq:video_correction} gives
\begin{equation}
\nabla_\theta\cL_{\mathrm{score}}
=-2\,\mathbb E\!\left[\frac{1}{d}J^\top\sg\!\left(\widetilde\dd_{t'}\right)\right].
\label{eq:video_replay_gradient}
\end{equation}
The factor $1/d$ follows from the mean-squared error. Rows of $J$ for the detached prefix are zero, while replay outputs retain derivatives through the connected causal representations inside the active segment. This is the exact gradient of the stated frozen-target surrogate.
For a one-step map $\mmG_\theta=\zz-\mmF_\theta$, the endpoint target $\xx_\theta+\widetilde\dd$ corresponds to the velocity target $\mmF_\theta-\widetilde\dd$. More generally, the clean-latent conversion at replay time $t_i>0$ maps an endpoint displacement $\widetilde\dd$ to a velocity-target displacement $-\widetilde\dd/t_i$.

\paragraph{Score interpretation.}
For $k\in\{r,\theta\}$, applying the OT conversion to a velocity estimate defines
\begin{equation}
\widehat\ss_{k,t'}(\yy\mid\cc)
:=-\frac{\yy+(1-t')\widehat\vv_{k,t'}(\yy\mid\cc)}{t'}.
\label{eq:apex_estimated_scores}
\end{equation}
Subtracting at the same state $\yy=\xx_{\theta,t'}^{\sg}$ gives \equref{eq:apex_score_difference}. A true conditional-score interpretation requires calibrated marginal velocity fields under common conditioning. The video update retains S2's teacher conditioning and guidance and uses the causal shared fake estimator of \secref{subsec:fake_branch}. These identities alone do not make the guided, normalized update an exact conditional or full-trajectory KL gradient.

\subsection{Endpoint Formulation}
\label{app:legacy_endpoint}

We retain the original APEX endpoint identities~\citep{liu2026self} below for their algebraic relation to velocity regression. These auxiliary losses are not terms in the video objective of \equref{eq:video_objective}. In this original shared-field construction only, the reference-role estimate $\widehat\vv_{r,t}$ is the original-condition call of $\mmF_\theta$; the video correction in \secref{sec:apex} instead uses the frozen teacher. The shifted-condition estimate remains $\widehat\vv_{\theta,t}$.

Here, the shifted field is learned from the time-dependent endpoint samples below; $p_{\theta,t}^{\mathrm{end}}$ denotes their re-noised marginal at the reused time $t$ under the same noise schedule as the reference samples.

This auxiliary endpoint construction forms a detached prediction from a partially noised reference sample:
\begin{equation}
\xx_\theta^{\mathrm{end}}(t):=\sg\!\left[\xx_t-t\mmF_{\theta}(\xx_t,t,\cc)\right].
\label{eq:one_step_fake}
\end{equation}

Reusing the same $t$, let $\xx_{\theta,t}^{\mathrm{end}}=(1-t)\xx_\theta^{\mathrm{end}}(t)+t\mepsilon'$ with independent $\mepsilon'\sim p_z$. Under $\cc_{\mathrm{shift}}=-\cc$, the auxiliary loss supervises the fake velocity in the shared network at this re-noised endpoint:
\begin{equation}
\cL_{\mathrm{fake}}^{\mathrm{end}}
:=\mathbb E\!\left[\left\|\mmF_{\theta}(\xx_{\theta,t}^{\mathrm{end}},t,\cc_{\mathrm{shift}})
-(\mepsilon'-\xx_\theta^{\mathrm{end}}(t))\right\|_2^2\right].
\label{eq:legacy_fake_loss}
\end{equation}

The superscript $\mathrm{end}$ distinguishes this auxiliary loss from the replay-based fake regression in \equref{eq:video_fake}. For endpoint regression, the branch evaluations at the same real-data path point are
\begin{equation}
\begin{aligned}
\widehat\vv_{r,t}&:=\mmF_{\theta}(\xx_t,t,\cc),\\
\widehat\vv_{\theta,t}&:=\sg\!\left[\mmF_{\theta}(\xx_t,t,\cc_{\mathrm{shift}})\right].
\end{aligned}
\label{eq:apex_v_def}
\end{equation}

Using the endpoint predictor in \equref{eq:endpoint_pred_app}, the mixed target and loss are
\begin{equation}
\mathbf T_{\mathrm{mix}}:=(1-\lambda)\xx+\lambda\mf^{\xx}(\widehat\vv_{\theta,t},\xx_t,t),
\qquad\lambda\in[0,1],
\label{eq:tmix}
\end{equation}
\begin{equation}
\cL_{\mathrm{mix}}:=\mathbb E\!\left[\frac{1}{\omega(t)}
\left\|\mf^{\xx}(\mmF_{\theta},\xx_t,t)-\mathbf T_{\mathrm{mix}}\right\|_2^2\right],
\qquad\omega(t)=\frac{t}{1-t}.
\label{eq:l_mix}
\end{equation}
The two component losses and their gradient-equivalent combination are
\begin{align}
\cL_{\mathrm{sup}}&:=\mathbb E\!\left[\frac{1}{\omega(t)}
\left\|\mf^{\xx}(\mmF_{\theta},\xx_t,t)-\xx\right\|_2^2\right],\label{eq:l_sup}\\
\cL_{\mathrm{cons}}&:=\mathbb E\!\left[\frac{1}{\omega(t)}
\left\|\mf^{\xx}(\mmF_{\theta},\xx_t,t)-\mf^{\xx}(\widehat\vv_{\theta,t},\xx_t,t)\right\|_2^2\right],\label{eq:l_cons}\\
\cG_{\text{\APEX}}&:=(1-\lambda)\cL_{\mathrm{sup}}+\lambda\cL_{\mathrm{cons}}.\label{eq:g_apex}
\end{align}
The mixing weight $\lambda$ and time weight $\omega(t)$ are used only in the following endpoint identities.

\subsection{Endpoint-Velocity Equivalence}
\label{app:endpoint-proof}

We prove that the endpoint-space MSE and velocity-space MSE are equivalent up to a scalar factor $t^2$.
Comparing weighted losses therefore requires accounting for this time-dependent factor.

\begin{proposition}[Endpoint-Velocity Equivalence for Supervised FM]
\label{prop:endpoint_velocity}

Let $\mf^{\xx}(\mmF, \xx_t, t) := \xx_t - t\,\mmF$ be the endpoint predictor defined in \equref{eq:endpoint_pred_app}, and let $\uu_r(\xx,\zz) = \zz - \xx$ be the target velocity under the OT path.

For any velocity estimate $\mmF_{\theta}$, the squared endpoint error is the squared velocity error scaled by $t^2$:
\begin{equation}
\bigl\|\mf^{\xx}(\mmF_{\theta}, \xx_t, t) - \xx\bigr\|_2^2
= t^2\,\bigl\|\mmF_{\theta} - \uu_r(\xx,\zz)\bigr\|_2^2.
\label{eq:endpoint_velocity_equiv1}
\end{equation}
\end{proposition}

\begin{proof}
\textbf{Step 1: Expand the endpoint predictor.}
\equref{eq:endpoint_pred_app} gives the endpoint prediction as
\begin{equation}
\mf^{\xx}(\mmF_{\theta}, \xx_t, t) = \xx_t - t\,\mmF_{\theta}.
\label{eq:endpoint_expand}
\end{equation}

\textbf{Step 2: Compute the squared error.}
The left-hand side of \equref{eq:endpoint_velocity_equiv1} is
\begin{align}
\bigl\|\mf^{\xx}(\mmF_{\theta}, \xx_t, t) - \xx\bigr\|_2^2
&= \bigl\|(\xx_t - t\,\mmF_{\theta}) - \xx\bigr\|_2^2 \nonumber \\
&= \bigl\|(\xx_t - \xx) - t\,\mmF_{\theta}\bigr\|_2^2.
\label{eq:endpoint_error_expand}
\end{align}

\textbf{Step 3: Use the OT path identity.}
Subtracting the clean endpoint from the OT path in \equref{eq:ot_path_app} gives
\begin{align}
\xx_t - \xx
&= \bigl[t\zz + (1-t)\xx\bigr] - \xx \quad \text{(substituting the OT path)} \nonumber \\
&= t\zz + (1-t)\xx - \xx \nonumber \\
&= t\zz + (1-t)\xx - 1 \cdot \xx \quad \text{(writing $\xx = 1 \cdot \xx$)} \nonumber \\
&= t\zz + \xx - t\xx - \xx \quad \text{(expanding $(1-t)\xx$)} \nonumber \\
&= t\zz - t\xx \quad \text{(canceling $\xx$)} \nonumber \\
&= t(\zz - \xx). \quad \text{(factoring out $t$)}
\label{eq:ot_diff}
\end{align}

Recall that under the OT path, the target velocity is defined as $\uu_r(\xx,\zz) := \zz - \xx$, which is the instantaneous rate of change from data $\xx$ to noise $\zz$.
Therefore, we obtain the key identity:
\begin{equation}
\xx_t - \xx = t\,\uu_r(\xx,\zz).
\label{eq:xt_minus_x}
\end{equation}
The target velocity is constant along this path because the interpolation is linear in time.

\textbf{Step 4: Substitute and simplify.}
Substituting \equref{eq:xt_minus_x} into \equref{eq:endpoint_error_expand}:
\begin{align}
\bigl\|(\xx_t - \xx) - t\,\mmF_{\theta}\bigr\|_2^2
&= \bigl\|t\,\uu_r(\xx,\zz) - t\,\mmF_{\theta}\bigr\|_2^2 \quad \text{using \equref{eq:xt_minus_x}} \nonumber \\
&= \bigl\|t\,(\uu_r(\xx,\zz) - \mmF_{\theta})\bigr\|_2^2 \quad \text{(factoring out $t$)} \nonumber \\
&= t^2\,\bigl\|\uu_r(\xx,\zz) - \mmF_{\theta}\bigr\|_2^2, \quad \text{(using $\|c\vv\|_2^2 = c^2\|\vv\|_2^2$)}
\label{eq:endpoint_final}
\end{align}
which proves \equref{eq:endpoint_velocity_equiv1}. The final step uses the homogeneity property of the squared $\ell_2$ norm.

\end{proof}

\begin{proposition}[Endpoint-Velocity Equivalence for Fake Alignment]
\label{prop:endpoint_velocity_fake}

For the fake-alignment term, let $\widehat\vv_{\theta,t}(\xx_t\mid\cc) := \sg\!\left(\mmF_{\theta}(\xx_t, t, \cc_{\mathrm{shift}})\right)$ be the fake velocity evaluated with the same online network $\mmF_{\theta}$ under the shifted condition $\cc_{\mathrm{shift}}$ with stop-gradient applied. Their squared endpoint difference is
\begin{equation}
\bigl\|\mf^{\xx}(\mmF_{\theta}, \xx_t, t) - \mf^{\xx}(\widehat\vv_{\theta,t}, \xx_t, t)\bigr\|_2^2
= t^2\,\bigl\|\mmF_{\theta} - \widehat\vv_{\theta,t}(\xx_t\mid\cc)\bigr\|_2^2.
\label{eq:endpoint_velocity_equiv2}
\end{equation}
\end{proposition}

\begin{proof}
\textbf{Step 1: Expand both endpoint predictors.}
By the definition of the endpoint predictor,
\begin{align}
\mf^{\xx}(\mmF_{\theta}, \xx_t, t) &= \xx_t - t\,\mmF_{\theta}, \\
\mf^{\xx}(\widehat\vv_{\theta,t}, \xx_t, t) &= \xx_t - t\,\widehat\vv_{\theta,t}(\xx_t\mid\cc).
\end{align}

\textbf{Step 2: Compute the difference.}
\begin{align}
\mf^{\xx}(\mmF_{\theta}, \xx_t, t) - \mf^{\xx}(\widehat\vv_{\theta,t}, \xx_t, t)
&= \bigl[\xx_t - t\,\mmF_{\theta}\bigr] - \bigl[\xx_t - t\,\widehat\vv_{\theta,t}\bigr] \nonumber \\
&= \xx_t - t\,\mmF_{\theta} - \xx_t + t\,\widehat\vv_{\theta,t} \nonumber \\
&= t\,\widehat\vv_{\theta,t} - t\,\mmF_{\theta} \nonumber \\
&= t\,(\widehat\vv_{\theta,t} - \mmF_{\theta}).
\label{eq:endpoint_diff_fake}
\end{align}

\textbf{Step 3: Square the norm.}
\begin{align}
\bigl\|\mf^{\xx}(\mmF_{\theta}, \xx_t, t) - \mf^{\xx}(\widehat\vv_{\theta,t}, \xx_t, t)\bigr\|_2^2
&= \bigl\|t\,(\widehat\vv_{\theta,t} - \mmF_{\theta})\bigr\|_2^2 \nonumber \\
&= t^2\,\bigl\|\widehat\vv_{\theta,t} - \mmF_{\theta}\bigr\|_2^2 \nonumber \\
&= t^2\,\bigl\|\mmF_{\theta} - \widehat\vv_{\theta,t}(\xx_t\mid\cc)\bigr\|_2^2,
\label{eq:endpoint_final_fake}
\end{align}
which proves \equref{eq:endpoint_velocity_equiv2} for both endpoint predictions at the same state and time.
\end{proof}

\subsection{Gradient Equivalence of Alternative Loss}
\label{app:alt-loss-proof}

We now show that the gradient of the mixed endpoint loss $\cL_{\text{mix}}$ is equal to the gradient of the alternative loss $\cG_{\text{\APEX}}$.
This equivalence holds when the mixed target is detached during differentiation.

\begin{theorem}[Gradient Equivalence]
\label{thm:gradient_equivalence}

Consider $\cL_{\text{mix}}$ from \equref{eq:l_mix} and $\cG_{\text{\APEX}}$ from \equref{eq:g_apex}.

Then, for any choice of parameter $\theta$, differentiating the two objectives gives
\begin{equation}
\nabla_{\theta}\,\cL_{\text{mix}}(\theta) = \nabla_{\theta}\,\cG_{\text{\APEX}}(\theta).
\label{eq:gradient_equiv_thm}
\end{equation}
\end{theorem}

\begin{proof}

We omit the expectation and work with one training pair, retaining the time weight $1/\omega(t)$.

Write $\mmF_\theta\equiv\mmF_\theta(\xx_t,t,\cc)$, $J_t:=\partial\mmF_\theta/\partial\theta\in\R^{d\times P}$, and $\widehat\vv_{\theta,t}\equiv\widehat\vv_{\theta,t}(\xx_t\mid\cc)$. The mixed target $\mathbf T_{\mathrm{mix}}$ includes the sampled clean data $\xx$. All target inputs stay fixed when differentiating the original-condition prediction.

\paragraph{Part A: Gradient of the mixed endpoint loss.}

\textbf{Step A1: Write the mixed endpoint loss.}

From \equref{eq:l_mix}, the mixed endpoint loss is written in terms of its detached target as
\begin{equation}
\cL_{\text{mix}}(\theta) = \frac{1}{\omega(t)}\bigl\|\mf^{\xx}(\mmF_{\theta}, \xx_t, t) - \mathbf T_{\mathrm{mix}}\bigr\|_2^2,
\label{eq:lmix_def}
\end{equation}

where the mixed target is defined in \equref{eq:tmix} as the following weighted sum:
\begin{equation}
\mathbf T_{\mathrm{mix}} = (1-\lambda)\,\xx + \lambda\,\mf^{\xx}(\widehat\vv_{\theta,t}, \xx_t, t).
\label{eq:tmix_def}
\end{equation}

\textbf{Step A2: Expand the endpoint predictors.}
Using the definition $\mf^{\xx}(\mmF, \xx_t, t) = \xx_t - t\,\mmF$ from \equref{eq:endpoint_pred_app}, the original and shifted predictions at the same noisy sample and time take the following forms:
\begin{align}
\mf^{\xx}(\mmF_{\theta}, \xx_t, t) &= \xx_t - t\,\mmF_{\theta}, \label{eq:fx_student} \\
\mf^{\xx}(\widehat\vv_{\theta,t}, \xx_t, t) &= \xx_t - t\,\widehat\vv_{\theta,t}. \label{eq:fx_fake}
\end{align}

\textbf{Step A3: Substitute into the mixed target.}
Substituting \equref{eq:fx_fake} into \equref{eq:tmix_def}:
\begin{align}
\mathbf T_{\mathrm{mix}}
&= (1-\lambda)\,\xx + \lambda\,(\xx_t - t\,\widehat\vv_{\theta,t}) \nonumber \\
&= (1-\lambda)\,\xx + \lambda\,\xx_t - \lambda t\,\widehat\vv_{\theta,t}.
\label{eq:tmix_expanded}
\end{align}

\textbf{Step A4: Compute the error term $\mdelta$.}
We define the endpoint residual by
\begin{equation}
\mdelta := \mf^{\xx}(\mmF_{\theta}, \xx_t, t) - \mathbf T_{\mathrm{mix}}.
\label{eq:delta_def}
\end{equation}
Substituting \equref{eq:fx_student} and \equref{eq:tmix_expanded} into this definition of the endpoint residual gives
\begin{align}
\mdelta
&= (\xx_t - t\,\mmF_{\theta}) - \bigl[(1-\lambda)\,\xx + \lambda\,\xx_t - \lambda t\,\widehat\vv_{\theta,t}\bigr] \nonumber \\
&= \xx_t - t\,\mmF_{\theta} - (1-\lambda)\,\xx - \lambda\,\xx_t + \lambda t\,\widehat\vv_{\theta,t} \nonumber \\
&= \xx_t(1 - \lambda) - (1-\lambda)\,\xx - t\,\mmF_{\theta} + \lambda t\,\widehat\vv_{\theta,t} \nonumber \\
&= (1-\lambda)(\xx_t - \xx) - t\,\mmF_{\theta} + \lambda t\,\widehat\vv_{\theta,t}.
\label{eq:delta_step1}
\end{align}

\textbf{Step A5: Apply the OT path identity.}
Using \equref{eq:xt_minus_x}, proved in \appref{app:endpoint-proof}, we obtain
\begin{equation}
\xx_t - \xx = t\,\uu_r, \quad \text{where } \uu_r = \zz - \xx.
\label{eq:ot_identity_use}
\end{equation}
Substituting into \equref{eq:delta_step1} expresses the endpoint residual through the velocity predictions:
\begin{align}
\mdelta
&= (1-\lambda)\,t\,\uu_r - t\,\mmF_{\theta} + \lambda t\,\widehat\vv_{\theta,t} \nonumber \\
&= t\bigl[(1-\lambda)\,\uu_r + \lambda\,\widehat\vv_{\theta,t} - \mmF_{\theta}\bigr].
\label{eq:delta_final}
\end{align}

\textbf{Step A6: Compute the gradient using the chain rule.}
The time weight is independent of $\theta$. For the vector residual $\mdelta\in\R^d$, the chain rule gives the following gradient with respect to $\theta$:
\begin{equation}
\nabla_\theta\cL_{\mathrm{mix}}
=\frac{2}{\omega(t)}\left(\frac{\partial\mdelta}{\partial\theta}\right)^{\!\top}\mdelta.
\label{eq:grad_lmix_chain}
\end{equation}
The sampled $\xx,\zz,t$ and the detached shifted prediction are held fixed during differentiation, so
\begin{equation}
\frac{\partial\mdelta}{\partial\theta}=-tJ_t.
\label{eq:grad_delta}
\end{equation}

\textbf{Step A7: Substitute the endpoint residual.}
Using \equref{eq:delta_final} and \equref{eq:grad_delta},
\begin{align}
\nabla_\theta\cL_{\mathrm{mix}}
&=\frac{2t^2}{\omega(t)}J_t^\top
\bigl[\mmF_\theta-(1-\lambda)\uu_r-\lambda\widehat\vv_{\theta,t}\bigr]\nonumber\\
&=2t(1-t)J_t^\top
\bigl[\mmF_\theta-(1-\lambda)\uu_r-\lambda\widehat\vv_{\theta,t}\bigr],
\label{eq:grad_lmix_step}
\end{align}
where $t^2/\omega(t)=t(1-t)$, as in \equref{eq:lweight_effective}, for the sampled time in this expectation.

\textbf{Step A8: Separate the two residuals.}
By linearity of multiplication by $J_t^\top$,
\begin{equation}
\nabla_\theta\cL_{\mathrm{mix}}
=2t(1-t)J_t^\top\bigl[(1-\lambda)(\mmF_\theta-\uu_r)
+\lambda(\mmF_\theta-\widehat\vv_{\theta,t})\bigr].
\label{eq:grad_lmix_final}
\end{equation}

\paragraph{Part B: Gradient of the alternative loss.}

\textbf{Step B1: Write the alternative loss.}

From \equref{eq:g_apex}, the alternative loss is the weighted combination of supervision and consistency:
\begin{equation}
\cG_{\text{\APEX}}(\theta) = (1-\lambda)\,\cL_{\text{sup}}(\theta) + \lambda\,\cL_{\text{cons}}(\theta),
\label{eq:gapex_def}
\end{equation}

where $\cL_{\text{sup}}$ and $\cL_{\text{cons}}$ are defined in \equref{eq:l_sup} and \equref{eq:l_cons}, respectively.

\textbf{Step B2: Apply the endpoint-velocity equivalence.}
By \propref{prop:endpoint_velocity}, we have
\begin{equation}
\cL_{\text{sup}}(\theta) = \frac{1}{\omega(t)}\bigl\|\mf^{\xx}(\mmF_{\theta}, \xx_t, t) - \xx\bigr\|_2^2
= \frac{t^2}{\omega(t)}\,\bigl\|\mmF_{\theta} - \uu_r\bigr\|_2^2.
\label{eq:lsup_velocity}
\end{equation}
By \propref{prop:endpoint_velocity_fake}, the fake-alignment term can be written as the following velocity error:
\begin{equation}
\cL_{\text{cons}}(\theta) = \frac{1}{\omega(t)}\bigl\|\mf^{\xx}(\mmF_{\theta}, \xx_t, t) - \mf^{\xx}(\widehat\vv_{\theta,t}, \xx_t, t)\bigr\|_2^2
= \frac{t^2}{\omega(t)}\,\bigl\|\mmF_{\theta} - \widehat\vv_{\theta,t}\bigr\|_2^2.
\label{eq:lcons_velocity}
\end{equation}

\textbf{Step B3: Compute the gradients of $\cL_{\text{sup}}$ and $\cL_{\text{cons}}$.}
Both losses carry the time weight $1/\omega(t)$ from \equref{eq:l_sup} and \equref{eq:l_cons}, in addition to $t^2$.
Since $\omega(t)=t/(1-t)$ depends only on the sampled time and not on $\xx_t$, the two factors combine into a single per-sample weight
\begin{equation}
\frac{1}{\omega(t)}\cdot t^2 = \frac{1-t}{t}\cdot t^2 = t\,(1-t),
\label{eq:lweight_effective}
\end{equation}
which remains explicit in both gradients and varies with the sampled time.

Applying the chain rule to the supervised and fake-alignment residuals gives
\begin{align}
\nabla_\theta\cL_{\mathrm{sup}}
&=2t(1-t)J_t^\top(\mmF_\theta-\uu_r),\label{eq:grad_lsup}\\
\nabla_\theta\cL_{\mathrm{cons}}
&=2t(1-t)J_t^\top(\mmF_\theta-\widehat\vv_{\theta,t}).\label{eq:grad_lcons}
\end{align}
The stop-gradient on $\widehat\vv_{\theta,t}$ blocks its derivative despite the shared parameters.

\textbf{Step B4: Combine the gradients.}
Substituting these expressions into \equref{eq:gapex_def} yields
\begin{align}
\nabla_\theta\cG_{\mathrm{APEX}}
&=(1-\lambda)\nabla_\theta\cL_{\mathrm{sup}}
+\lambda\nabla_\theta\cL_{\mathrm{cons}}\nonumber\\
&=2t(1-t)J_t^\top\bigl[(1-\lambda)(\mmF_\theta-\uu_r)
+\lambda(\mmF_\theta-\widehat\vv_{\theta,t})\bigr].
\label{eq:grad_gapex_final}
\end{align}

\paragraph{Part C: Conclusion.}

Comparing \equref{eq:grad_lmix_final} and \equref{eq:grad_gapex_final}, we see that the gradients match:
\begin{equation}
\nabla_{\theta}\,\cL_{\text{mix}}(\theta) = \nabla_{\theta}\,\cG_{\text{\APEX}}(\theta).
\end{equation}

The equality holds for each sampled pair. Taking expectations under the same sampling law proves \thmref{thm:gradient_equivalence}, assuming differentiation with respect to $\theta$ and expectation can be interchanged.
\end{proof}

\subsection{Score-Difference Form of the Endpoint Surrogate Gradient}
\label{app:score_diff_form}

All expectations below use real-data pairs $(\xx,\cc)$, independent noise $\zz\sim p_z$, and the sampled time $t\in(0,1)$; $\xx_t=(1-t)\xx+t\zz$.
The shifted-condition prediction is detached within the endpoint loss, so only the original-condition prediction contributes a derivative with respect to the network parameters $\theta$.

\begin{proposition}[Score-Difference Form]
\label{prop:score_diff_form}
Let $\widehat\vv_{r,t}$ and $\widehat\vv_{\theta,t}$ be the original- and shifted-condition predictions from \appref{app:legacy_endpoint}, evaluated at $(\xx_t,\cc)$.
Define their induced fields by $\widehat\ss_{i,t}:=-[\xx_t+(1-t)\widehat\vv_{i,t}]/t$ for $i\in\{r,\theta\}$, and let $\ss_{r,t}=\nabla_{\xx_t}\log p_{r,t}(\xx_t\mid\cc)$ be the exact noised-data score.
The target field combines the true data score and the induced fake score with the endpoint mixing weight $\lambda$:
\begin{equation}
\widehat\ss_{\mathrm{mix},t}:=(1-\lambda)\ss_{r,t}+\lambda\widehat\ss_{\theta,t}.
\label{eq:score_mix_def}
\end{equation}
For $\omega(t)=t/(1-t)$ and $J_t:=J_\theta\mmF_\theta(\xx_t,t,\cc)$, the frozen-target gradient is
\begin{equation}
\nabla_\theta\cG_{\mathrm{APEX}}
=-2\mathbb E\!\left[t^2J_t^\top\sg\!\left(\widehat\ss_{r,t}-\widehat\ss_{\mathrm{mix},t}\right)\right].
\label{eq:apex_score_diff_grad}
\end{equation}
The coefficient $-2t^2$ depends on time, but is independent of the sample at a fixed $t$.
\end{proposition}

\begin{proof}
Because $J_t$ depends on the sampled pair only through $(\xx_t,t,\cc)$, conditional expectation replaces the supervised target $\uu_r=\zz-\xx$ with the true data velocity $\vv_r^*(\xx_t,t,\cc)$.
The field entering the gradient of $\cG_{\mathrm{APEX}}=(1-\lambda)\cL_{\mathrm{sup}}+\lambda\cL_{\mathrm{cons}}$ is the following weighted combination of velocity residuals:
\begin{equation}
(1-\lambda)(\widehat\vv_{r,t}-\vv_r^*)+\lambda(\widehat\vv_{r,t}-\widehat\vv_{\theta,t}).
\label{eq:velocity_combo}
\end{equation}
The OT conversion and the definitions of the induced fields give the following two identities:
\begin{align}
\widehat\vv_{r,t}-\vv_r^*
&=-\frac{t}{1-t}(\widehat\ss_{r,t}-\ss_{r,t}),\label{eq:vel_score_1}\\
\widehat\vv_{r,t}-\widehat\vv_{\theta,t}
&=-\frac{t}{1-t}(\widehat\ss_{r,t}-\widehat\ss_{\theta,t}).\label{eq:vel_score_2}
\end{align}
Combining these identities expresses the velocity residual in score coordinates:
\begin{equation}
(1-\lambda)(\widehat\vv_{r,t}-\vv_r^*)+\lambda(\widehat\vv_{r,t}-\widehat\vv_{\theta,t})
=-\frac{t}{1-t}(\widehat\ss_{r,t}-\widehat\ss_{\mathrm{mix},t}).
\label{eq:velocity_to_score_mix}
\end{equation}
Substituting this identity into the weighted endpoint gradient with respect to $\theta$ yields
\begin{align}
\nabla_\theta\cG_{\mathrm{APEX}}
&=\mathbb E\!\left[\frac{2t^2}{\omega(t)}J_t^\top\sg\!\left((1-\lambda)(\widehat\vv_{r,t}-\vv_r^*)+\lambda(\widehat\vv_{r,t}-\widehat\vv_{\theta,t})\right)\right]\nonumber\\
&=-2\mathbb E\!\left[t^2J_t^\top\sg\!\left(\widehat\ss_{r,t}-\widehat\ss_{\mathrm{mix},t}\right)\right],
\label{eq:grad_apex_score}
\end{align}
where $t^2/\omega(t)=t(1-t)$.
All time-dependent factors remain inside the expectation over $t$.
\end{proof}

\begin{remark}[Scope of the gradient identity]
\label{rem:score_diff_scope}
This identity rewrites the frozen-target gradient of the endpoint objective.
It does not require the network-induced fields $\widehat\ss_{r,t}$ and $\widehat\ss_{\theta,t}$ to equal true distribution scores, and it does not identify the loss with a divergence between generated and data distributions.
The coefficient is constant across samples only at a fixed time; the residual and network Jacobian still vary across samples.
\end{remark}

\begin{remark}[When the mixed field has a density interpretation]
\label{rem:score_mix_law}

For each fixed $t$, $p_{\theta,t}^{\mathrm{end}}$ is the law of $(1-t)\xx_\theta^{\mathrm{end}}(t)+t\mepsilon'$ from \appref{app:legacy_endpoint}, with independent Gaussian $\mepsilon'$. The clean endpoint law itself depends on $t$; the score conversion applies at each fixed time.
If the shifted-condition field is calibrated so that $\widehat\ss_{\theta,t}=\nabla\log p_{\theta,t}^{\mathrm{end}}$, then $\widehat\ss_{\mathrm{mix},t}$ is the score of the following geometric combination:
\begin{equation}
p_{\mathrm{mix},t}(\xx_t\mid\cc)\propto
p_{r,t}(\xx_t\mid\cc)^{1-\lambda}
\left[p_{\theta,t}^{\mathrm{end}}(\xx_t\mid\cc)\right]^\lambda,
\label{eq:pmix_geometric}
\end{equation}
provided this expression is normalizable.
This is a geometric combination at a fixed time, and need not define a consistent probability path across time.
Without calibration, $\widehat\ss_{\mathrm{mix},t}$ need not be a density score.
\end{remark}

\paragraph{Relation to GAN weighting.}

For comparison, define the noised generated sample $\xx_{\theta,t}:=(1-t)\mmG_\theta(\zz,\cc)+t\mepsilon'$ and its Jacobian $J_t^G:=J_\theta\xx_{\theta,t}$. At fixed noise, the non-saturating GAN gradient takes the form
\begin{equation}
\nabla_\theta\cL_{\mathrm{NS\text{-}GAN}}
=\mathbb E\!\left[(J_t^G)^\top\,w_{\mathrm{NS}}(\xx_{\theta,t})
\left(\ss_{\theta,t}(\xx_{\theta,t})-\ss_{r,t}(\xx_{\theta,t})\right)\right],
\label{eq:gan_ns_grad}
\end{equation}
where $w_{\mathrm{NS}}=1-D_t^*$ for the optimal discriminator held fixed during the generator update.
Unlike this discriminator-dependent weight, the explicit coefficient in \equref{eq:apex_score_diff_grad} depends only on time.
Each gradient uses its own expectation and Jacobian; this comparison concerns sample weighting.
\subsection{GAN Dynamics and Score-Difference Gradients}
\label{sec:gan-dynamics}
Let $p_{\theta}(\xx)$ and $p_r(\xx)$ denote the generated and data densities for a fixed condition $\cc$, omitted below for brevity, and let $\ss(\xx):=\nabla_{\xx}\log p(\xx)$ denote a distribution's score.
With the optimal discriminator $D^*(\xx)=\nicefrac{p_r(\xx)}{p_r(\xx)+p_{\theta}(\xx)}$ held fixed during the generator update~\citep{mohamed2016learning,goodfellow2014generative}, the saturating and non-saturating losses yield a weighted score-difference gradient:
\begin{equation}
  \nabla_{\theta} \cL_{\text{GAN}} \propto \EEb{\xx \sim p_{\theta}}{ \left(\frac{\partial \xx}{\partial \theta}\right)^{\!\top} w_{\mathrm{GAN}}(\xx)\bigl(\ss_{\theta}(\xx)-\ss_r(\xx)\bigr) } \,.
  \label{eq:gan_grad}
\end{equation}
For $\xx=\mmG_{\theta}(\zz,\cc)$, the weights are $w_{\mathrm{GAN}}(\xx)=D^*(\xx)$ (saturating) and $w_{\mathrm{GAN}}(\xx)=1-D^*(\xx)$ (non-saturating).

%% file: sections/8_training_data.tex
\clearpage
\section{Training Data}
\label{sec:data}
We reuse the complete S2-Avatar training data and preparation pipeline, including filtering and annotation~\citep{zhang2026vidu}.
The data include livestream and talking-head videos, film and television, solo dance, and 2D/3D animation, covering facial expressions, body motion, and animated characters.
\figref{fig:data_pipeline} summarizes the data pipeline inherited from Vidu S1~\citep{vidus1}.

Single-shot clips are filtered for subject visibility, visual quality, content safety, shot stability, and interactivity.
Smooth camera motion is corrected using background features after masking foreground subjects; clips with large or complex camera motion are excluded.
Full-clip and event-aligned chunk captions describe actions and outcomes chronologically, using object detection and speech recognition to ground the annotations.
The clipping, quality checks, background stabilization, and annotation procedures follow S2 without changes; APEX modifies only the model training procedure, as described in \secref{sec:video_method}.

\input{figures/8_data_pipeline.tex}

%% file: figures/8_data_pipeline.tex
\begin{figure}[!htb]
\centering
\begingroup
\fontsize{9}{11}\selectfont
\begin{minipage}[t]{0.42\linewidth}
\centering
{\fontsize{10}{12}\selectfont\textbf{Five-stage pipeline}}\par\medskip
\fbox{\parbox{0.86\linewidth}{\centering Clipping}}\\[2pt]
$\downarrow$\\[2pt]
\fbox{\parbox{0.86\linewidth}{\centering Filtering}}\\[2pt]
$\downarrow$\\[2pt]
\fbox{\parbox{0.86\linewidth}{\centering Speech Processing}}\\[2pt]
$\downarrow$\\[2pt]
\fbox{\parbox{0.86\linewidth}{\centering Captioning}}\\[2pt]
$\downarrow$\\[2pt]
\fbox{\parbox{0.86\linewidth}{\centering Embedding}}
\end{minipage}\hfill
\begin{minipage}[t]{0.52\linewidth}
{\fontsize{10}{12}\selectfont\textbf{S2-Avatar processing}}\par\medskip
A vision-language model checks candidate cuts for subtle edits.\par\medskip
Filtering selects clear videos and stabilizes suitable backgrounds.\par\medskip
Captions describe full clips and event-based chunks, grounded in objects and recognized speech.\par\medskip
APEX uses the same S2-Avatar data and processing.
\end{minipage}
\endgroup
\caption{\textbf{S2-Avatar data preparation}, summarized from Vidu S2~\citep{zhang2026vidu}. Left: the five-stage pipeline reused in our experiments. Right: the processing steps summarized in \protect\appref{sec:data}.}
\label{fig:data_pipeline}
\end{figure}

%% file: sections/9_evaluation_protocol.tex
\newpage
\section{Evaluation Details}
\label{app:video_protocol}
We detail the evaluators, aggregation rules, and comparison budgets for \secref{sec:video_results} and~\secref{sec:video_additional}.

\subsection{StreamAV-Bench}
\label{sec:video_public}
We evaluate all 320 StreamAV-Bench cases~\citep{liu2026streamavbench}: 160 Progressive and 160 Interactive scenarios, using 180-second outputs.
The Interactive track introduces five runtime condition updates at 30-second intervals, at 30, 60, 90, 120, and 150 seconds.
We measure APEX and S2-Avatar. Other models' VIF scores are from StreamAV-Bench~\citep{liu2026streamavbench}, with other scores from Vidu S2~\citep{zhang2026vidu}.

Visual Aesthetics (VA) uses the LAION Aesthetic Predictor~\citep{laion2022aesthetic}.
Visual Quality (VQ), Video Instruction Fulfillment (VIF), Audio Quality (AQ), and Audio Instruction Fulfillment (AIF) use the benchmark's Gemini judge.
Subject Consistency (SC) and Background Consistency (BC) measure full-sequence stability.
Audio-Visual Alignment (AVAlign) uses ImageBind similarity~\citep{girdhar2023imagebind} to assess cross-modal semantic agreement.
Audio-Visual Synchronization (AVSync) uses Synchformer~\citep{iashin2024synchformer} to measure temporal offset error in seconds.
Production Quality (PQ) assesses the output audio with AudioBox Aesthetics~\citep{tjandra2025audiobox}.
All benchmark metrics except AVSync are higher-is-better.

VA, PQ, AVAlign, and AVSync are evaluated over six non-overlapping 30-second windows per video, whereas SC and BC use the full 180-second sequence.
Track summaries average case-level scores with equal weight. Overall summaries pool all 320 cases. Since the tracks have equal size, this equals the average of the two unrounded track means.
The 12-model comparison in \tabref{tab:streamav_overall} uses Progressive cases for VIF, AIF, SC, and BC, and both tracks for the remaining metrics.
\tabref{tab:streamav_tracks} provides track means and pooled results. Differences use unrounded APEX minus S2-Avatar scores, so a positive AVSync difference indicates worse synchronization.

\subsection{Paired Human Comparisons}
Each comparison uses 70 paired 180-second I2V cases.
The five preference dimensions are consistency, instruction following, audiovisual synchronization, video quality, and motion quality.
Good/Same/Bad (GSB) counts wins, ties, and losses for the evaluated model relative to the reference, with $G+S+B=70$ for each dimension.
Net preference is $100(G-B)/70$, retaining tied cases in the denominator. One net win therefore changes the reported preference by approximately 1.43 percentage points.
Mean Opinion Scores (MOS) assess instruction following and synchronization, with higher values indicating better quality.
A sequence is counted as usable when both scores exceed 2.
Synchronization MOS and the benchmark's AVSync error therefore have opposite directions.
In the main study, consistency and video quality each have G/S/B counts of 13/47/10 (three net wins), whereas motion quality has 2/60/8 (six net losses), matching \figref{fig:video_preference}.

The main 3-NFE human study in \figref{fig:video_preference} and the additional sampling and training comparisons use separate evaluation rounds. \tabref{tab:training_comparisons} groups the latter by sampling budget and optimizer configuration.
Each additional variant is paired with S2-Avatar at 3 NFE, so differences between variants are comparisons against a common reference rather than direct pairwise judgments.
For the optimizer comparisons, shared AdamW states are evaluated at 3 NFE and separate states at 2 NFE, with model parameters shared in both settings.
All replay-time sampling variants share optimizer states and sort sampled times along the sequence. At noise levels $(0.75,0.9375,0.99)$, the default, DF4, and DF10 weights correspond to sampling probabilities $(1/3,1/3,1/3)$, $(2/3,1/6,1/6)$, and $(5/6,1/12,1/12)$, respectively. DF10 favors the lowest noise level most strongly.

\subsection{Backbone and Refiner Budget}
NFE counts backbone denoising evaluations per generated video block, excluding KV-cache encoding and updates.
All paired comparisons use the same one-step spatial Refiner with $2\times$ upsampling, following S2~\citep{zhang2026vidu}.
The Refiner adds one separate evaluation per block to either a 2-NFE or a 3-NFE backbone.
The 2-NFE setting therefore uses one third fewer backbone evaluations, with the Refiner unchanged.
The teacher and fake estimator add no inference calls.
The replay times $t_i$ and scoring time $t'$ in \secref{sec:video_method} are variables used during training and do not determine the inference evaluation budget.